\documentclass[runningheads]{llncs}

\usepackage{eccv}

\usepackage{eccvabbrv}

\usepackage{graphicx}
\usepackage{booktabs}
\usepackage{multirow}
\usepackage[normalem]{ulem}
\useunder{\uline}{\ul}{}

\usepackage{algorithm}
\usepackage{algorithmic}

\usepackage[accsupp]{axessibility}  

\usepackage[breaklinks,colorlinks,citecolor=eccvblue,linkcolor=eccvblue,urlcolor=eccvblue]{hyperref}

\usepackage{orcidlink}

\begin{document}
	
	\title{From Reconstruction to Decision: A Post-Encoder Plug-in Adapter for Curvilinear Segmentation}
	
	\titlerunning{A Post-Encoder Plug-in Adapter for Curvilinear Segmentation}
	
	\author{
		Qin Lei\inst{1,2}\orcidlink{0000-0002-1340-5969}
		\and
		Jiang Zhong\inst{4}
		\and
		Xin Xiao\inst{4}
		\and
		Yuming Yang\inst{4}
		\and
		Hao Wu\inst{1,2,3}\orcidlink{0009-0001-9116-1066}\thanks{Corresponding author.}
	}
	
	\authorrunning{Q. Lei et al.}
	
	\institute{
		Center for Big Data and Intelligent Medicine, The First Affiliated Hospital of Chongqing Medical University, Chongqing, China
		\and
		Key Laboratory of Digital Health and Intelligent Medicine, Chongqing Municipal Health Commission, Chongqing, China
		\and
		Chongqing Translational Medicine Center, Chongqing, China
		\and
		College of Computer Science, Chongqing University, Chongqing, China\\
		\texttt{qinlei@hospital.cqmu.edu.cn};
		\texttt{zhongjiang@cqu.edu.cn};\\
		\texttt{20241401023@stu.cqu.edu.cn};
		\texttt{ymyang@cqu.edu.cn};
		\texttt{wuhao@cqmu.edu.cn};
	}
	
	\maketitle

	\begin{abstract}
		Curvilinear object segmentation, including vessels and cracks, is challenging due to extreme spatial sparsity and topological fragility, where small local errors can cause severe structural disconnections. Meanwhile, modern segmentation pipelines increasingly rely on strong but hard-to-modify foundation encoders whose heavy downsampling limits fine structural recovery. Motivated by this, we focus on the post-encoder stage and study two recurring and actionable failure modes: a reconstruction bottleneck in high-resolution feature restoration and a decision bottleneck in binarization. We present PEPA, a lightweight Post-Encoder Plug-in Adapter for 2D curvilinear segmentation pipelines with accessible decoder/head features and target, query, or class descriptors. PEPA couples (i) Target-Conditioned Snake Upsampling (TCSU), which uses target-conditioned continuous snake-like sampling to better recover thin and tortuous structures during upsampling, and (ii) Target-Adaptive Differentiable Thresholding (TADT), which predicts target-specific thresholds and optimizes a soft-threshold surrogate with explicit safeguards against trivial bias shifting. Under this post-encoder interface, PEPA can be attached to both prompt-based decoders and conventional dense predictors. Experiments on five medical and industrial benchmarks show that adding PEPA to frozen-encoder baselines yields consistent improvements, with gains in topological connectivity (clDice) typically exceeding those in region overlap (IoU), indicating improved structural continuity. With only $\sim$0.26M additional parameters, PEPA offers a practical post-encoder enhancement for structure-centric segmentation.
		
	\end{abstract}

\section{Introduction}
\label{sec:intro}

Segmenting curvilinear objects, such as vessels~\cite{mou2021cs2,haft2019deep}, neuronal branches~\cite{shit2021cldice,liu2024dneuromat}, and cracks~\cite{lei2024integrating,liu2019deepcrack,chen2023devil,chen2024mind}, is a core challenge in computer vision. 
Unlike compact macro-objects, curvilinear structures exhibit extreme spatial sparsity and topological fragility~\cite{lei2025enhancing,shit2021cldice,chen2025self}. 
Their thin profiles and drastic local contrast decay make their topological integrity highly sensitive to high-frequency detail reconstruction and the final binarization boundary~\cite{lei2024joint,lei2023dynamic,lei2023adaptive}. 
Consequently, even minor pixel-level errors can cause severe structural disconnections and undermine downstream analyses~\cite{shit2021cldice,lei2024expanding,lei2024enriching}.

Vision Foundation Models (VFMs), such as the SAM series~\cite{kirillov2023segment,ravi2024sam} and DINO series~\cite{oquab2023dinov2,simeoni2025dinov3}, have recently provided strong visual representations for dense prediction. 
However, applying them to curvilinear structures exposes a \textit{semantic-spatial paradox}: foundation encoders obtain semantic abstraction through heavy downsampling (e.g., $16\times$), but this process also removes the fine spatial granularity required to delineate delicate topologies. 
Since full fine-tuning of massive encoders is computationally expensive and may compromise their general representations, structure preservation is often delegated to the post-encoder stage. 
This raises a practical question: \textit{How can we design a lightweight post-encoder plug-in that improves the reconstruction and final decision of fragile curvilinear structures without modifying the foundation encoder?}

\begin{figure}[htpb]
	\centering
	\includegraphics[width=\textwidth]{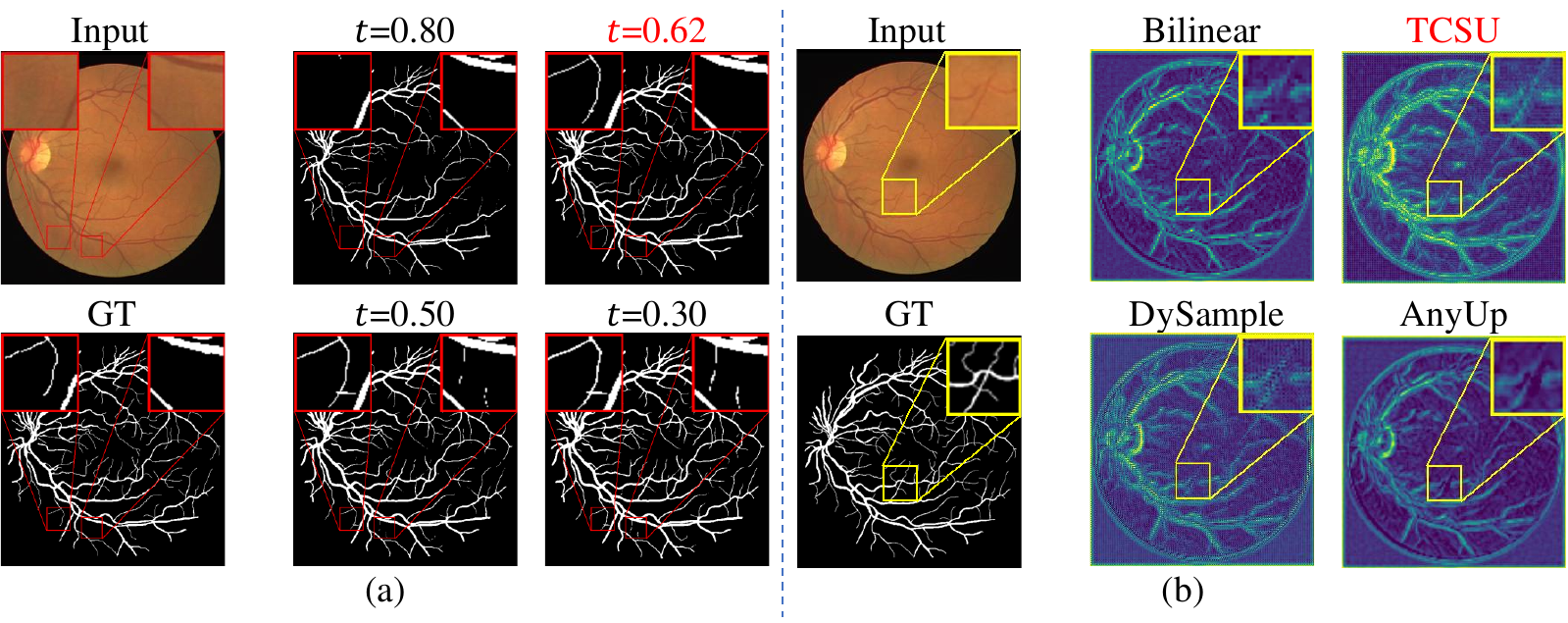}
	\caption{Motivation for our proposed PEPA. \textbf{(a) Decision Bottleneck (TADT):} Applying a static global threshold ($t=0.80, 0.50, 0.30$) either breaks faint branches or introduces severe noise. A target-adaptive threshold ($t=0.62$, red) balances connectivity and noise suppression. \textbf{(b) Reconstruction Bottleneck (TCSU):} Traditional bilinear interpolation and recent advanced upsamplers, such as DySample~\cite{liu2023learning} and AnyUp~\cite{wimmer2025anyup}, produce blurred or disconnected features for thin vessels. Our Target-Conditioned Snake Upsampling (TCSU) synthesizes sharper and more continuous structural responses.}
	\label{fig:motivation}
\end{figure}

We focus on two recurring and actionable post-encoder bottlenecks for curvilinear segmentation: the \textbf{Reconstruction Bottleneck} of high-resolution feature restoration and the \textbf{Decision Bottleneck} of probability discretization. 
These bottlenecks are not intended to exhaust all possible failure modes; rather, they characterize two common stages where fragile structures are frequently broken after low-resolution semantic embeddings have been produced by the encoder.

In the reconstruction phase, decoders typically rely on isotropic, \textit{target-agnostic} operations such as bilinear upsampling. 
As shown in Fig.~\ref{fig:motivation}(b), passive interpolation can blur thin responses and cause topology breakage. 
Even recent advanced upsamplers, such as AnyUp~\cite{wimmer2025anyup} and DySample~\cite{liu2023learning}, are not explicitly constrained by curvilinear morphology, so strong gradients from thick structures may dominate the reconstruction process while faint branches remain fragmented. 
For topology-sensitive targets, upsampling should therefore move beyond passive interpolation toward active, morphology-conditioned geometric reconstruction.

In the decision phase, converting continuous probabilities into binary masks is sensitive to the threshold choice. 
As shown in Fig.~\ref{fig:motivation}(a), a high global threshold can suppress faint terminal branches, whereas a low threshold can introduce noisy false positives. 
Although differentiable binarization has been effective in scene-level tasks such as text detection~\cite{liao2022real}, na\"{\i}vely applying learnable thresholds to dense, multi-target curvilinear segmentation can lead to \textit{trivial bias compensation}: the network may jointly shift logits and thresholds to bypass true boundary improvement rather than learning a more topology-preserving decision boundary.

To address these two post-encoder bottlenecks, we propose \textbf{PEPA (Post-Encoder Plug-in Adapter)}, a modular and lightweight adapter for 2D curvilinear segmentation pipelines with accessible decoder/head features and target, query, or class descriptors. 
PEPA shifts the post-encoder stage from rigid reconstruction and fixed discretization to target-adaptive reconstruction and decision calibration through two core modules:
\begin{itemize}
	\item \textbf{Target-Conditioned Snake Upsampling (TCSU):} TCSU generates sub-pixel sampling points along continuous and dynamically sized snake-like neighborhoods. By modulating chain length and deformation direction with a target descriptor, TCSU encourages high-resolution reconstruction to follow the queried structure while reducing interference from distractor gradients.
	\item \textbf{Target-Adaptive Differentiable Thresholding (TADT):} TADT predicts target-specific binarization thresholds and constructs an end-to-end optimizable soft-threshold surrogate. To mitigate threshold-learning degeneration, TADT uses logit centering, topology-aware supervision on the surrogate, and local threshold-perturbation consistency.
\end{itemize}

In summary, our main contributions are three-fold:
\begin{itemize}
	\item We identify reconstruction and decision calibration as two recurring post-encoder bottlenecks for curvilinear segmentation, and propose \textbf{TCSU}, a target-conditioned upsampling operator that uses continuous snake-like sampling chains to synthesize structure-preserving high-resolution features.
	\item We design \textbf{TADT}, a target-adaptive differentiable thresholding framework that optimizes the soft binarization surrogate with topology-aware objectives and explicit anti-degeneration safeguards.
	\item We integrate TCSU and TADT into \textbf{PEPA}, a lightweight post-encoder plug-in for 2D curvilinear segmentation pipelines. Extensive experiments across five medical and industrial benchmarks demonstrate that equipping frozen Vision Foundation Models with PEPA yields consistent improvements, achieving average absolute gains of \textbf{+2.6\% in IoU and +2.8\% in clDice} without fine-tuning the massive backbone encoders.
\end{itemize}

\section{Related Work}

\subsection{Feature Upsampling}
In some visual tasks, feature upsampling is essential for restoring high-resolution spatial details from low-resolution semantic embeddings. Traditional methods typically rely on task-agnostic operations, such as bilinear interpolation or standard transposed convolutions, which process all spatial dimensions isotropically and often cause topological drift or edge blurring in thin structures~\cite{liu2023learning}. 
To address this, learnable upsamplers like CARAFE~\cite{wang2019carafe} and DySample~\cite{liu2023learning} introduce dynamic, content-aware sampling strategies.
Recently, with the rise of Vision Foundation Models (VFMs), arbitrary-resolution and feature-agnostic upsamplers have emerged, including FeatUp~\cite{fu2024featup}, LoftUp~\cite{huang2025loftup}, JAFAR~\cite{couairon2025jafar}, and AnyUp~\cite{wimmer2025anyup}.
While these methods achieve state-of-the-art performance in general semantic segmentation, they apply globally-attended reconstruction without explicit morphological constraints. Consequently, the attention mechanisms are often hijacked by the gradients of thick background structures, leading to disconnected features for faint curvilinear targets. 
In contrast, our Target-Conditioned Snake Upsampling (TCSU) transitions from passive interpolation to active, morphology-conditioned geometric deformation, explicitly tracking the queried object's topology.

\subsection{Differentiable Binarization and Adaptive Thresholding}
Converting continuous probability maps into discrete binary masks typically relies on rigid global thresholds, which struggle to balance noise suppression and connectivity preservation due to the high intra-class variance of curvilinear structures~\cite{lei2024joint}. 
To enable end-to-end optimization of the binarization process, DBNet~\cite{liao2020real, liao2022real} proposed a differentiable approximate step function, achieving great success in scene text detection. 
Recently, this concept has been broadly extended to various vision domains; 
for instance, BAA~\cite{shu2025binarization} incorporates binarization behavior into gradient-based optimization for edge detection, while BI-DiffSR~\cite{chen2024binarized} and BiMaCoSR~\cite{liu2025bimacosr} utilize customized binarization architectures to compress diffusion models for image super-resolution.
In the domain of crack and curvilinear segmentation, Lei et al.~\cite{lei2023dynamic, lei2024joint, lei2024integrating} advanced this paradigm by reformulating the task as a multi-objective problem, jointly optimizing dynamic pixel-level thresholds and utilizing causal augmentation to mitigate extreme class imbalance. 
Despite these advancements, directly applying differentiable binarization to dense, multi-target curvilinear segmentation often triggers trivial bias compensation, where the network jointly shifts logits and thresholds rather than sharpening actual decision boundaries. 
Our Target-Adaptive Differentiable Thresholding (TADT) module evades this degradation via logit centering and direct calibration with connectivity-aware losses.

\section{Method}
\label{sec:method}

Curvilinear targets exhibit extreme spatial sparsity and structural fragility, making their topological integrity highly sensitive to high-resolution reconstruction and the final binary decision. Since modern segmentation increasingly relies on frozen foundation encoders, improvements must be \emph{plug-and-play} across architectures. We propose \textbf{PEPA} (\textbf{P}ost-\textbf{E}ncoder \textbf{P}lug-in \textbf{A}dapter), coupling \textbf{Target-Conditioned Snake Upsampling (TCSU)} for structure-preserving reconstruction with \textbf{Target-Adaptive Differentiable Thresholding (TADT)} for decision calibration.

\paragraph{Notation.}
Given encoder features $\mathbf{F}\in\mathbb{R}^{B\times C\times H\times W}$ and a target descriptor $\mathbf{e}_k\in\mathbb{R}^{B\times d}$ (e.g., a prompted object or class embedding), TCSU upsamples $\mathbf{F}$ to a high-resolution substrate $\mathbf{U}_k\in\mathbb{R}^{B\times C'\times sH\times sW}$ (scale $s$). The mask head then outputs logits $\mathbf{z}_k$. Concurrently, TADT predicts a target-adaptive threshold $t_k$ from $\mathbf{e}_k$, enabling differentiable binarization during training and hard thresholding at inference.

\subsection{Target-Conditioned Snake Upsampling (TCSU)}
\label{sec:tcsu}

TCSU synthesizes high-resolution features by sampling along \emph{snake-shaped neighborhoods}. Unlike standard interpolation, TCSU employs adaptive range and continuity-constrained deformation, both modulated by the target descriptor $\mathbf{e}_k$ to track specific curvilinear topologies while avoiding clutter. 

\begin{figure}[t]
	\centering
	\includegraphics[width=\linewidth]{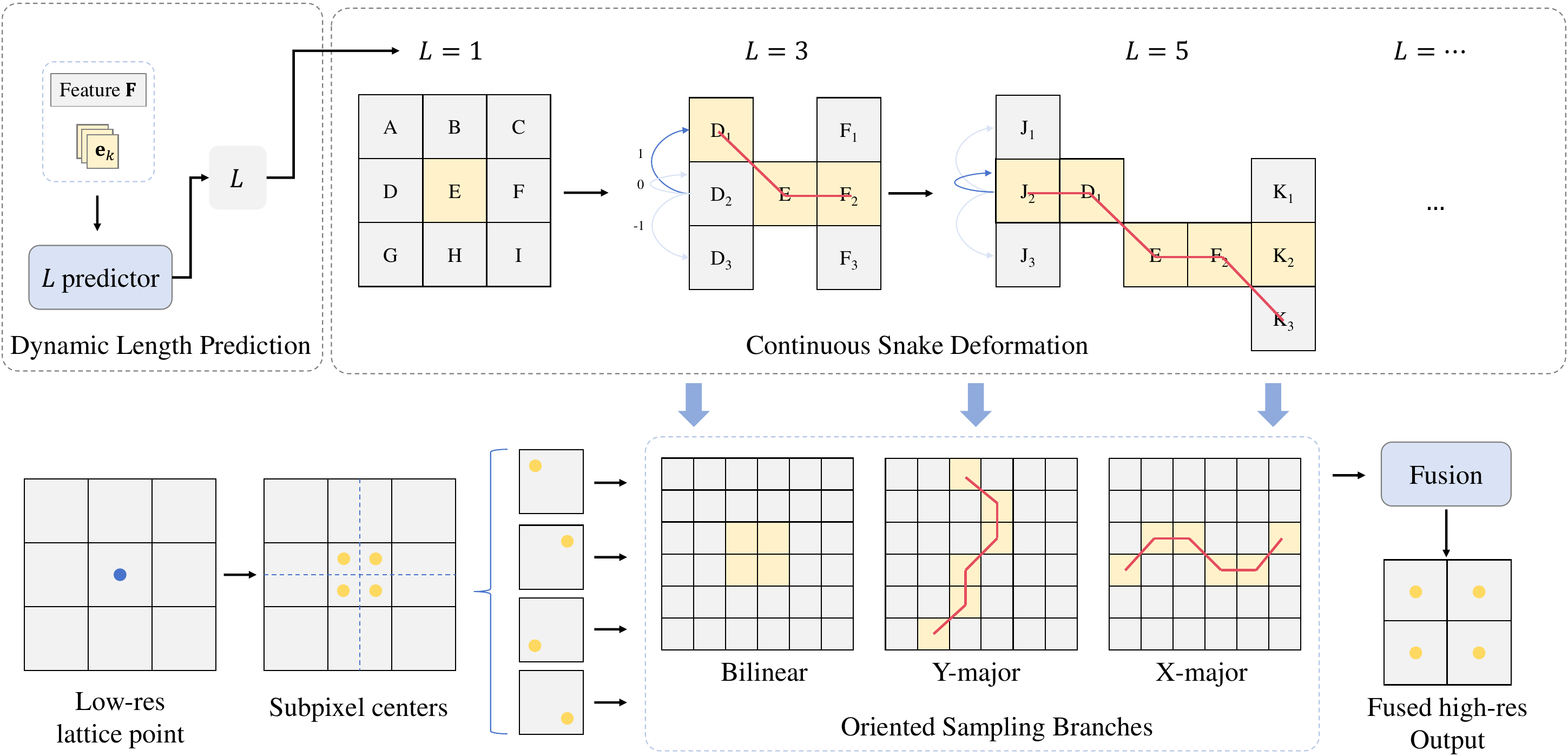}
	\caption{\textbf{TCSU architecture.} The module predicts a dynamic sampling length $L$ and accumulates continuous snake deformations. The features are sampled via $X$-major and $Y$-major branches, and then fused with a bilinear shortcut to construct the final high-resolution output.}
	\label{fig:tcsu_diagram}
\end{figure}

Taking $2\times$ upsampling ($s{=}2$) as an example, each low-resolution lattice $(x,y)$ introduces four subpixel centers $\mathbf{p}^{0}_{m}(x,y) = (x,y) + \boldsymbol{\pi}_m$. Alongside a stable bilinear shortcut $\mathbf{U}^{\text{bi}}_k = \mathrm{BilinearUp}(\mathbf{F}; s)$, TCSU branches reconstruct features via three sequential operations:

\noindent\textbf{1) Dynamic Length Prediction.} We predict an effective chain length $L_k$ to control the snake's extent. We modulate a base length $L_{\text{base}}$ using feature evidence $\mathrm{Summ}(\mathbf{F})$ and the target $\mathbf{e}_k$, applying a straight-through estimator to enforce a symmetric odd length:
\begin{equation}
	L_k = \mathrm{OddRound}\Big(L_{\text{base}}\cdot \big(1 + g(\mathrm{Summ}(\mathbf{F}), \mathbf{e}_k)\big)\Big) \in \{1,3,5,\dots\}.
\end{equation}

\noindent\textbf{2) Conditioned Deformation \& Sampling.} We instantiate an X-major snake (bending in $y$) and a Y-major snake (bending in $x$). For subpixel $m$, incremental offsets are predicted via a shared-plus-refinement network: $f_{\star}(\mathbf{F},\mathbf{e}_k;m) = f_{\star,\text{sh}}(\mathbf{F};m) + \eta_k f_{\star,\text{tg}}(\mathbf{F},\mathbf{e}_k;m)$, where $\eta_k = \sigma(\mathrm{MLP}(\mathbf{e}_k))$. A smooth length-aware mask $\omega(i;L_k)$ truncates the increments to keep the dynamic length coupling differentiable: $\Delta y^{x}_{k,m,i} \leftarrow \Delta y^{x}_{k,m,i}\cdot \omega(i;L_k)$.

To preserve the structural adjacency prior of curvilinear objects, we iteratively accumulate these increments from the center outward. For the X-major snake, which extends along the $x$-axis while bending in $y$, the accumulated vertical offsets are defined as:
\begin{equation}
	\tilde{\Delta y}^{x}_{k,m}(0)=0,\qquad
	\tilde{\Delta y}^{x}_{k,m}(\pm i)=\tilde{\Delta y}^{x}_{k,m}(\pm (i-1)) + \Delta y^{x}_{k,m,\pm i},\quad i=1,\dots,c,
\end{equation}
where $c = \lfloor (K-1)/2 \rfloor$ and $K$ is the maximal chain length. Features are then differentiably sampled at these continuous coordinates:
\begin{equation}
	\mathbf{V}^{x}_{k,m} = \mathcal{S}\!\left(\mathbf{F}, \big\{\big(x + \Delta x^{0}_m + i,\; y + \Delta y^{0}_m + \tilde{\Delta y}^{x}_{k,m}(i)\big)\big\}_{i=-c}^{c}\right),
\end{equation}
with an analogous operation for $\mathbf{V}^{y}_{k,m}$. 

\noindent\textbf{3) Aggregation \& Fusion.} The sampled tensors are aggregated using 1D depthwise convolutions $\mathcal{A}_{\star}$ modulated by $L_k$. The subpixel responses $\mathbf{u}^{x}_{k,m}$ and $\mathbf{u}^{y}_{k,m}$ are rearranged onto the high-resolution grid and concatenated with the bilinear shortcut to yield the final $\mathbf{U}_k$.

\textbf{Relation to Snake Convolution.} Unlike standard snake convolutions~\cite{qi2023dynamic} which act as target-agnostic feature extractors, TCSU is a \emph{reconstruction} operator that explicitly conditions both its dynamic length and deformation on the target $\mathbf{e}_k$, enabling distinct structural recovery behaviors for different queries.

\subsection{Target-Adaptive Differentiable Thresholding (TADT)}
\label{sec:tadt}

A global threshold (e.g., 0.5) often forces a sub-optimal trade-off between breaking faint branches and introducing noise. TADT overcomes this by calibrating target-specific decision boundaries, predicting a bounded threshold $t_k \in [t_{\min}, t_{\max}]$ in the logit domain:
\begin{equation}
	t_k = t_{\min} + (t_{\max}-t_{\min})\cdot \sigma\!\big(\tilde{h}_t(\mathbf{e}_k,\mathrm{Summ}(\mathbf{F}))\big).
\end{equation}

To enable end-to-end optimization, we replace the non-differentiable step function with a smooth surrogate $\mathbf{b}_k = \sigma(\alpha(\tilde{\mathbf{z}}_k - t_k))$, where the scalar $\alpha$ controls the sharpness of the transition. In our implementation, $\alpha$ is empirically set to a constant value (e.g., $\alpha{=}1.0$) to maintain gradient stability without requiring complex annealing schedules. 

Crucially, jointly learning thresholds and logits often degenerates into trivial bias shifting. TADT prevents this via three mechanisms:
\textbf{(i) Logit Centering:} We remove the spatial mean $\tilde{\mathbf{z}}_k = \mathbf{z}_k - \mu(\mathbf{z}_k)$ to isolate relative confidence.
\textbf{(ii) Threshold-Aware Objectives:} Losses (e.g., clDice~\cite{shit2021cldice}) are computed directly on the surrogate $\mathbf{b}_k$, forcing $t_k$ to optimize topological connectivity.
\textbf{(iii) Local Stability:} To prevent the decision boundary from becoming overly sensitive to localized noise, we compute two perturbed surrogates by shifting the threshold by a margin $\Delta$ (set to 0.5 in the logit domain):
\begin{equation}
	\mathbf{b}_k^{\pm} = \sigma\!\left(\alpha\big(\tilde{\mathbf{z}}_k-(t_k\pm\Delta)\big)\right).
\end{equation}
We then enforce consistency between these perturbed states using a soft Dice agreement loss, explicitly formulated as:
\begin{equation}
	\mathcal{L}_{\text{consist}}(\mathbf{b}_k^{+},\mathbf{b}_k^{-}) = 1 - \frac{2 \sum \mathbf{b}_k^{+} \cdot \mathbf{b}_k^{-} + \epsilon}{\sum \mathbf{b}_k^{+} + \sum \mathbf{b}_k^{-} + \epsilon},
\end{equation}
which firmly anchors the dynamic boundary and discourages noise-driven fluctuations.

\subsection{Instantiations and Optimization}

\begin{figure}[htpb]
	\centering
	\includegraphics[width=\textwidth]{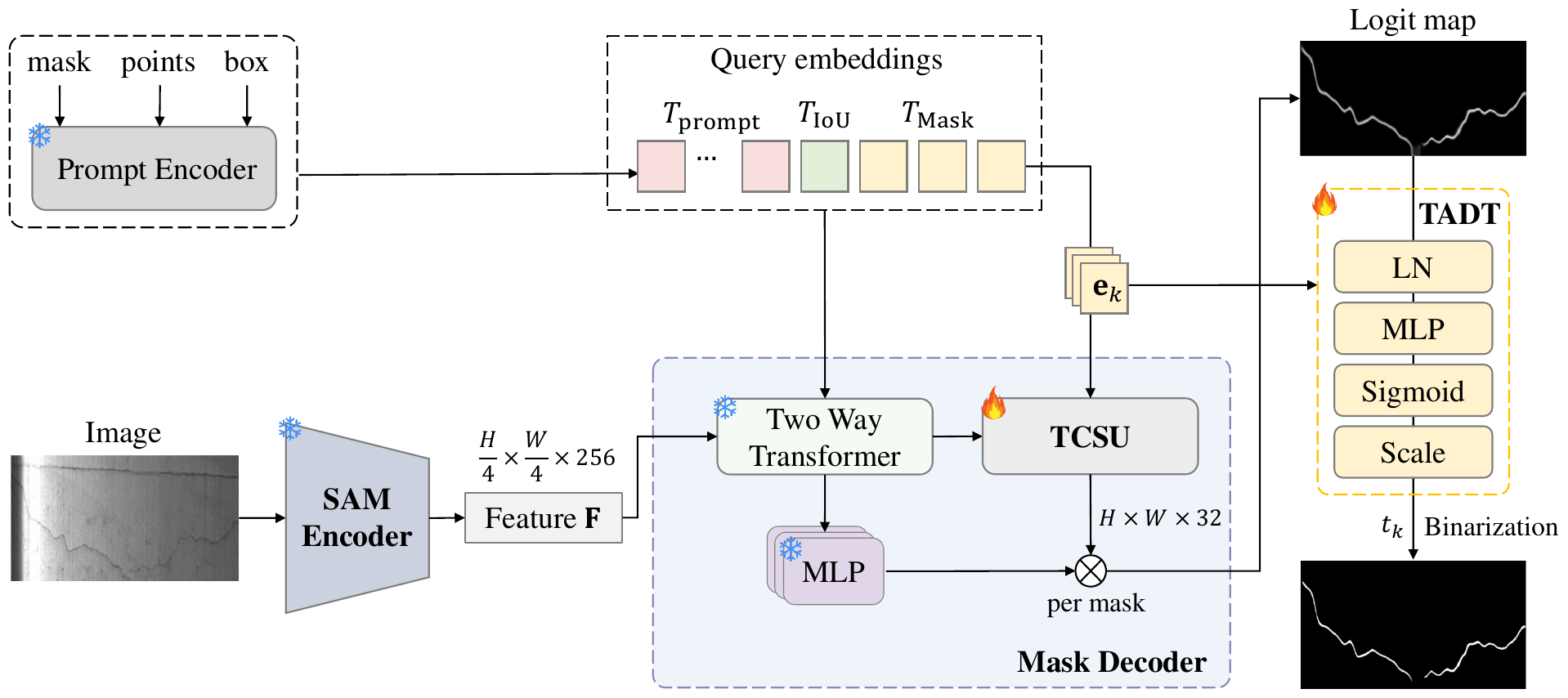}
	\caption{\textbf{PEPA SAM instantiation.} TCSU replaces the standard upsampling substrate, generating target-conditioned high-resolution features from query embeddings. Concurrently, TADT predicts a specific binarization threshold for the target.}
	\label{fig:pepa_sam}
\end{figure}

PEPA is designed as a lightweight post-encoder plug-in for 2D curvilinear segmentation pipelines with accessible decoder/head features and target, query, or class descriptors. In prompt-based models (e.g., SAM, as shown in Fig.~\ref{fig:pepa_sam}), $\mathbf{e}_k$ is the output mask token, and TCSU replaces the decoder's standard upscaling. 
In semantic models (e.g., U-Net), $\mathbf{e}_k$ is a learnable class embedding, and TCSU modules replace all hierarchical upsampling layers. 
During training, we optimize a composite loss evaluated on the binarization surrogate; when multiple mask hypotheses are produced, we choose $k^\star$ by minimizing this loss, otherwise no hypothesis selection is used.
\begin{equation}
	\mathcal{L}_{\text{total}} = \lambda_{\text{bce}}\mathcal{L}_{\text{bce}} + \lambda_{\text{dice}}\mathcal{L}_{\text{dice}} + \lambda_{\text{cl}}\big(1-\mathrm{clDice}\big) + \lambda_{\text{consist}}\mathcal{L}_{\text{consist}}.
\end{equation}

Detailed mathematical formulations of the subpixel initialization and network hyper-parameters are provided in the Supplementary Material.

\section{Experiments}

\subsection{Experimental Settings}
To evaluate the generalization of PEPA across diverse scenarios, we conduct experiments on five curvilinear segmentation benchmarks spanning medical vasculature and industrial scenes: DRIVE~\cite{staal2004ridge}, CHASEDB1~\cite{fraz2012ensemble}, CHUAC~\cite{cervantes2019automatic}, XCAD~\cite{ma2021self}, and Crack500~\cite{yang2019feature}. 
For quantitative evaluation, we employ Intersection over Union (IoU) to assess region-level classification accuracy and centerline Dice (clDice)~\cite{shit2021cldice} to measure topological connectivity and structural integrity.

\textbf{Implementation Details.} 
For the Vision Foundation Model experiments, we utilized the ViT-B backbone for all SAM variants (SAM, SAM-HQ, MedSAM) and the EfficientSAM-S backbone for EfficientSAM. During training, the massive foundation encoders were kept strictly frozen, and only the original mask decoders along with the PEPA modules were updated. Optimization was performed using the AdamW optimizer with an initial learning rate of $1 \times 10^{-4}$, a weight decay of $1 \times 10^{-4}$, and a cosine annealing scheduler for 100 epochs on a single NVIDIA H200 GPU. The batch size was set to 8 for all datasets. 

\textbf{Prompt Protocol.}
For prompt-based models, we follow the same prompt simulation strategy as SAM-HQ~\cite{ke2023segment} during training.
Given the ground-truth (GT) mask, we construct three candidate prompts: 
(i) a tight bounding box computed from the GT mask; 
(ii) $k{=}10$ positive point prompts uniformly sampled from foreground pixels; and 
(iii) a noisy mask prompt obtained by downsampling the GT mask to $256{\times}256$ and injecting random perturbations, which is fed to SAM as \texttt{mask\_inputs}.
For each training sample, we randomly select \emph{one} prompt type from \{\texttt{box}, \texttt{point}, \texttt{mask\_inputs}\}.
If the foreground region contains fewer than $k$ pixels (so that point sampling is ill-defined), we disable point prompts and sample from \{\texttt{box}, \texttt{mask\_inputs}\} only.
Following SAM-HQ, we use the \emph{single-mask} output mode (i.e., \texttt{multimask\_output=False}), hence no oracle selection among multiple mask hypotheses ($k^{*}$) is involved.

At test time, we adopt a strict box-prompt protocol: \emph{all} quantitative results are produced using only the GT-derived bounding box, without any additional prompts.
For qualitative visualization in Fig.~\ref{fig:exp_visual}, we additionally show results under manually chosen point and box prompts to better reflect interactive use cases.

\begin{table}[htpb]
	\centering
	\caption{\textbf{Decoder fine-tuning with frozen VFM encoders.} For all baseline models (w/o PEPA), their original segmentation decoders were fully fine-tuned. We compare these optimized baselines against the integration of PEPA (+PEPA) under the identical training protocol.}
	\label{tab:vfm_decoder_ft_pepa}
	\small
	\resizebox{\columnwidth}{!}{%
		\begin{tabular}{@{}lcccccccccccc@{}}
			\toprule
			\multirow{2}{*}{Method} &
			\multicolumn{2}{c}{XCAD} &
			\multicolumn{2}{c}{CHUAC} &
			\multicolumn{2}{c}{DRIVE} &
			\multicolumn{2}{c}{CHASEDB1} &
			\multicolumn{2}{c}{Crack500} &
			\multicolumn{2}{c}{Avg.} \\
			\cmidrule(lr){2-13}
			& IoU & clDice & IoU & clDice & IoU & clDice & IoU & clDice & IoU & clDice & IoU & clDice \\
			\midrule
			
			SAM\cite{kirillov2023segment}                   & 68.4 & 81.5 & 65.8 & 79.9 & 70.2 & 81.0 & 66.8 & 79.9 & 63.4 & 77.2 & 66.9 & 79.9 \\
			\quad + PEPA          & 73.1 & 85.3 & 67.8 & 81.1 & 72.8 & 84.2 & 70.8 & 85.5 & 64.8 & 79.6 & 69.9 & 83.1 \\
			\quad $\Delta \uparrow$ & +4.7 & +3.8 & +2.0 & +1.2 & +2.6 & +3.2 & +4.0 & +5.6 & +1.4 & +2.4 & +2.9 & +3.2 \\
			\midrule
			
			SAM-HQ\cite{ke2023segment}                & 68.7 & 81.8 & 66.2 & 79.2 & 70.0 & 80.8 & 66.3 & 79.4 & 63.8 & 77.6 & 67.0 & 79.8 \\
			\quad + PEPA          & 73.3 & 85.5 & 68.0 & 81.5 & 71.9 & 83.8 & 70.5 & 85.1 & 65.0 & 79.8 & 69.7 & 83.1 \\
			\quad $\Delta \uparrow$ & +4.6 & +3.7 & +1.8 & +2.3 & +1.9 & +3.0 & +4.2 & +5.7 & +1.2 & +2.2 & +2.7 & +3.4 \\
			\midrule
			
			MedSAM\cite{ma2024segment}                 & 68.6 & 81.7 & 65.4 & 79.5 & 70.5 & 81.3 & 66.4 & 79.6 & 63.5 & 77.8 & 66.9 & 80.0 \\
			\quad + PEPA          & 73.1 & 85.4 & 67.6 & 81.0 & 72.6 & 83.6 & 70.8 & 85.6 & 64.9 & 79.6 & 69.8 & 83.0 \\
			\quad $\Delta \uparrow$ & +4.5 & +3.7 & +2.2 & +1.5 & +2.1 & +2.3 & +4.4 & +6.0 & +1.4 & +1.8 & +2.9 & +3.1 \\
			\midrule
			
			EfficientSAM\cite{xiong2024efficientsam}          & 65.7 & 79.2 & 63.1 & 77.4 & 68.6 & 77.6 & 63.6 & 77.3 & 59.9 & 72.4 & 64.2 & 76.8 \\
			\quad + PEPA          & 69.2 & 81.7 & 65.3 & 79.2 & 70.3 & 81.5 & 68.4 & 81.4 & 62.6 & 77.4 & 67.2 & 80.2 \\
			\quad $\Delta \uparrow$ & +3.5 & +2.5 & +2.2 & +1.8 & +1.7 & +3.9 & +4.8 & +4.1 & +2.7 & +5.0 & +3.0 & +3.5 \\
			\midrule
			
			DINOv3(Mask2Former)\cite{cheng2022masked}  & 67.0 & 80.2 & 63.8 & 77.9 & 67.9 & 80.9 & 67.9 & 80.9 & 61.7 & 76.3 & 65.7 & 79.2 \\
			\quad + PEPA          & 69.1 & 82.1 & 65.8 & 79.8 & 69.7 & 82.5 & 70.0 & 82.8 & 63.6 & 78.2 & 67.6 & 81.1 \\
			\quad $\Delta \uparrow$ & +2.1 & +1.9 & +2.0 & +1.9 & +1.8 & +1.6 & +2.1 & +1.9 & +1.9 & +1.9 & +2.0 & +1.8 \\
			\midrule
			
			DINOv3(MaskDINO)\cite{li2023mask}     & 68.1 & 80.9 & 65.0 & 78.7 & 69.5 & 81.9 & 67.8 & 80.7 & 62.0 & 76.4 & 66.5 & 79.7 \\
			\quad + PEPA          & 70.5 & 83.0 & 67.1 & 80.6 & 71.4 & 83.6 & 70.1 & 82.7 & 63.9 & 78.3 & 68.6 & 81.6 \\
			\quad $\Delta \uparrow$ & +2.4 & +2.1 & +2.1 & +1.9 & +1.9 & +1.7 & +2.3 & +2.0 & +1.9 & +1.9 & +2.1 & +1.9 \\
			\midrule
			
			Avg.\ $\Delta \uparrow$ & +3.7 & +3.0 & +2.1 & +1.8 & +2.0 & +2.6 & +3.6 & +4.2 & +1.8 & +2.5 & +2.6 & +2.8 \\
			\bottomrule
		\end{tabular}%
	}
\end{table}

\subsection{Experimental Results}

\subsubsection{Enhancing Vision Foundation Models}

To demonstrate PEPA's plug-and-play capability, we integrate it into six Vision Foundation Models (VFMs), encompassing prompt-based architectures (e.g., SAM variants~\cite{kirillov2023segment,ke2023segment,ma2024segment,xiong2024efficientsam}) and conventional segmentation heads (e.g., DINOv3~\cite{simeoni2025dinov3} with Mask2Former~\cite{cheng2022masked} and MaskDINO~\cite{li2023mask}). Crucially, to ensure a fair comparison, the original decoders of all baseline models (w/o PEPA) were fully fine-tuned on the respective datasets. 

As shown in Table~\ref{tab:vfm_decoder_ft_pepa}, the consistent improvements yielded by PEPA demonstrate architectural gains strictly \textit{on top of} already optimized baselines. Specifically, PEPA achieves average absolute increases of \textbf{+2.6\% in IoU} and \textbf{+2.8\% in clDice}. A critical observation from these results is the metric asymmetry: the improvements in clDice are consistently more pronounced than those in IoU. This empirical evidence validates our core hypothesis that PEPA effectively addresses topological fragility rather than merely inflating pixel-wise overlap. 
\begin{figure}[htpb]
	\centering
	\includegraphics[width=\columnwidth]{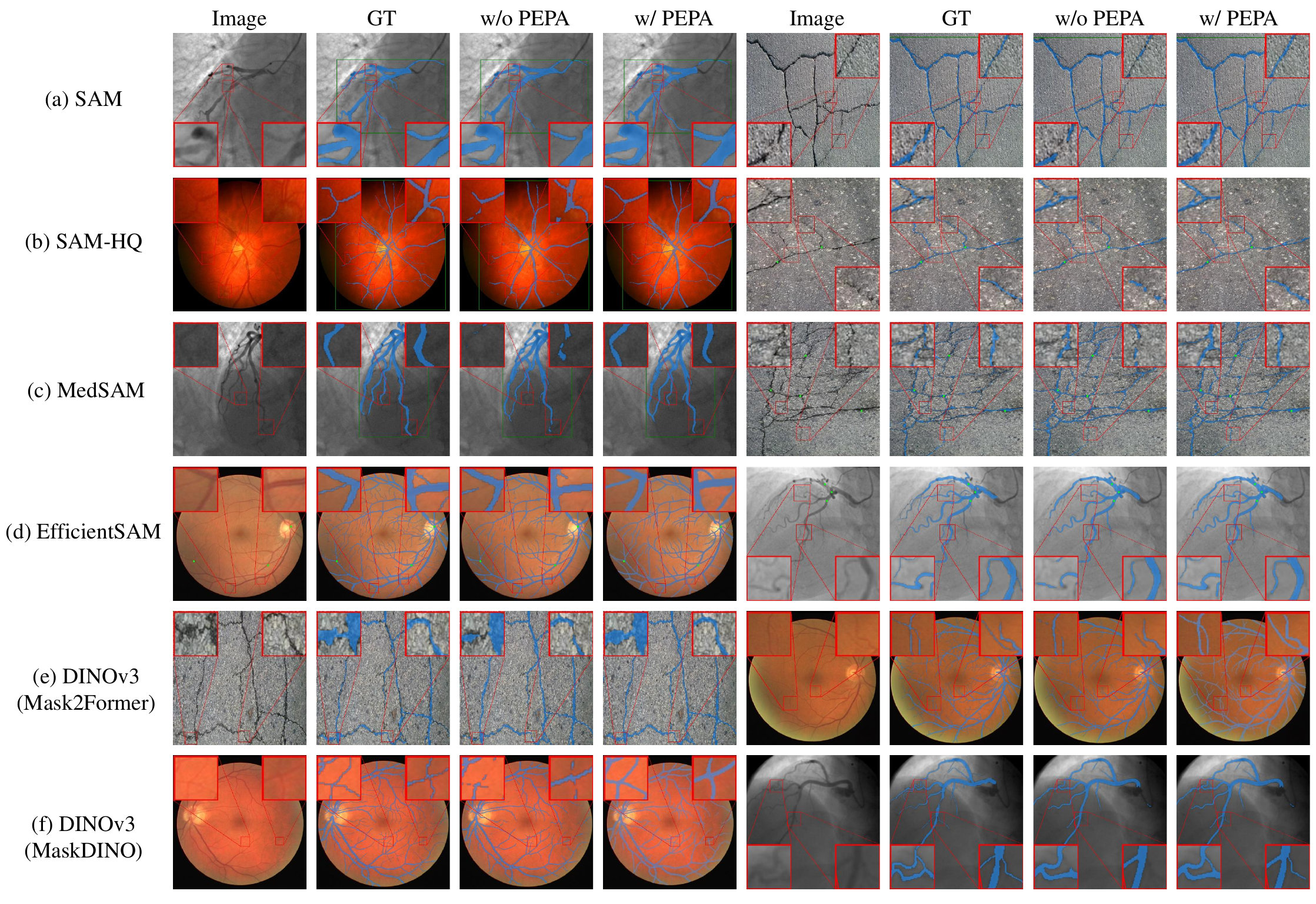}
	\caption{\textbf{Qualitative comparison of VFMs with and without PEPA.} Green markers (points and bounding boxes) indicate the spatial prompts provided to interactive models. Red boxes highlight zoomed-in local regions placed at the corners, demonstrating PEPA's superior ability to restore fragile branches and preserve continuous curvilinear structures.}
	\label{fig:exp_visual}
\end{figure}
Notably, prompt-based models experience massive topological gains on datasets characterized by extremely thin structures and low contrast. For instance, on the CHASEDB1 dataset, adding PEPA to SAM and MedSAM yields clDice gains of +5.6\% and +6.0\%, respectively. Qualitatively, while the fine-tuned baseline decoders still suffer from severe disconnections and noise at terminal vessel branches, PEPA-equipped models successfully bridge structural gaps and cleanly delineate faint networks (Fig.~\ref{fig:exp_visual}). 

\subsubsection{Comparison with Recent Domain-Specific Models}
We further evaluate our best-performing variant, \textbf{PEPA SAM}, against recent domain-specific networks. These include models specialized for crack detection (CrossDiff~\cite{shi2025crossdiff}, DBCNet~\cite{zhang2025dual}), coronary angiography (TVS-Net~\cite{he2025deep}, Mid-Net~\cite{zhao2025mid}), and retinal vessels (HM-Mamba~\cite{wang2025hierarchical}, GCC-UNet~\cite{wei2024retinal}), alongside the strong fully-convolutional nnU-Net~\cite{isensee2021nnu} and the prompt-based FPBE SAM~\cite{lei2025enhancing} baselines.

Despite keeping the massive image encoder entirely frozen, PEPA SAM consistently outperforms full-parameter fine-tuned domain experts and advanced prompt-based adapters across all datasets (Table~\ref{tab:my-table}). For instance, on the challenging XCAD dataset, PEPA SAM surpasses the highly competitive TVS-Net and Mid-Net, achieving an outstanding clDice of 85.3\%. Crucially, when compared directly to FPBE SAM—a recent state-of-the-art adapter specifically designed to enhance SAM for curvilinear structures—PEPA SAM demonstrates a distinct superiority in preserving structural continuity. While FPBE SAM achieves highly competitive region-level overlap (often securing the second-best IoU), PEPA SAM consistently outperforms it in topological metrics, yielding absolute clDice improvements of +2.2\% on XCAD and +2.3\% on CHASEDB1. 

Similarly, on retinal datasets, PEPA SAM outperforms HM-Mamba, a recent architecture utilizing State-Space Models (SSMs) for long-range dependency modeling. While SSMs excel at capturing global context, they still succumb to the continuous-to-discrete decision bottleneck during final binarization. PEPA circumvents this by calibrating thresholds adaptively, underscoring that overcoming the reconstruction-decision bottleneck at the decoding stage is a highly effective strategy for structure-preserving segmentation across diverse application domains.

\begin{table}[htpb]
	\centering
	\caption{\textbf{Quantitative comparison against recent domain-specific segmentation models.} Metrics are IoU and clDice. The best results are highlighted in \textbf{bold}, and the second best are \underline{underlined}.}
\label{tab:my-table}
	\resizebox{0.85\columnwidth}{!}{%
		\begin{tabular}{@{}lcccccccccc@{}}
			\toprule
			\multirow{2}{*}{Method} & \multicolumn{2}{c}{XCAD} & \multicolumn{2}{c}{CHUAC} & \multicolumn{2}{c}{DRIVE} & \multicolumn{2}{c}{CHASEDB1} & \multicolumn{2}{c}{Crack500} \\ \cmidrule(l){2-11} 
			& IoU & clDice & IoU & clDice & IoU & clDice & IoU & clDice & IoU & clDice \\ \midrule
			nnU-Net\cite{isensee2021nnu} & 70.6 & 81.8 & 66.5 & 78.6 & 70.4 & 81.2 & 67.6 & 78.7 & 63.1 & 76.2 \\
			CrossDiff\cite{shi2025crossdiff} & 65.0 & 77.2 & 62.5 & 74.3 & 66.0 & 78.0 & 63.4 & 76.0 & 64.0 & 78.5 \\
			DBCNet\cite{zhang2025dual} & 65.7 & 78.0 & 63.2 & 75.0 & 66.6 & 78.4 & 64.2 & 75.8 & {\ul 64.7} & {\ul 78.8} \\
			TVS-Net\cite{he2025deep} & 71.4 & {\ul 83.2} & 66.8 & 78.1 & 69.2 & 81.1 & 66.5 & 78.0 & 61.0 & 72.0 \\
			Mid-Net\cite{zhao2025mid} & 70.5 & 82.3 & 66.2 & {\ul 79.4} & 68.4 & 80.3 & 66.1 & 78.8 & 60.7 & 74.8 \\
			HM-Mamba\cite{wang2025hierarchical} & 66.8 & 78.5 & 64.3 & 76.0 & 71.8 & {\ul 83.2} & 68.9 & 81.5 & 60.4 & 72.0 \\
			GCC-UNet\cite{wei2024retinal} & 64.5 & 82.0 & 62.8 & 76.2 & 71.4 & 82.5 & 68.3 & 80.5 & 58.8 & 73.5 \\
			FPBE SAM~\cite{lei2025enhancing} & {\ul 72.2} & 83.1 & {\ul 67.3} & 79.2 & {\ul 72.1} & 83.1 & {\ul 70.3} & {\ul 83.2} & 64.1 & 78.6 \\ \midrule
			\textbf{PEPA SAM} & \textbf{73.1} & \textbf{85.3} & \textbf{67.8} & \textbf{81.1} & \textbf{72.8} & \textbf{84.2} & \textbf{70.8} & \textbf{85.5} & \textbf{64.8} & \textbf{79.6} \\ \bottomrule
		\end{tabular}%
	}
\end{table}

\subsubsection{Ablation Studies}
To validate the individual contributions and synergy of our proposed modules, we conduct extensive ablations on both medical and industrial benchmarks.

\noindent\textbf{Core Components.} As analyzed in Table~\ref{tab:ablation_xcad_crack500}, integrating either TCSU or TADT independently into conventional (nnU-Net) or prompt-based (SAM) baselines yields consistent gains. Interestingly, TADT provides a more significant clDice boost for SAM compared to nnU-Net. This is because SAM's default global thresholding was initially pre-trained on natural macro-objects, rendering it highly miscalibrated for fragile micro-structures. TCSU effectively mitigates the reconstruction bottleneck by actively tracking spatial morphologies, while TADT calibrates these decision boundaries to rescue faint responses. Crucially, their integration is strictly complementary, achieving peak performance and confirming that these dual bottlenecks must be decoupled and resolved jointly.

\begin{table}[htpb]
	\centering
	\caption{\textbf{Core module ablation on conventional (nnU-Net) and prompt-based (SAM) architectures.} The combination of TCSU and TADT demonstrates strictly complementary improvements.}
	\label{tab:ablation_xcad_crack500}
	\resizebox{0.7\columnwidth}{!}{%
		\begin{tabular}{@{}lcccccc@{}}
			\toprule
			\multirow{2}{*}{Baseline} & \multirow{2}{*}{TCSU} & \multirow{2}{*}{TADT} & \multicolumn{2}{c}{XCAD} & \multicolumn{2}{c}{Crack500} \\ \cmidrule(l){4-7} 
			&  &  & IoU $\uparrow$ & clDice $\uparrow$ & IoU $\uparrow$ & clDice $\uparrow$ \\ \midrule
			nnU-Net &  &  & 70.6 & 81.8 & 63.1 & 76.2 \\
			nnU-Net & $\checkmark$ &  & 71.2 & 82.6 & 63.8 & 76.9 \\
			nnU-Net &  & $\checkmark$ & 71.1 & 82.7 & 63.4 & 77.2 \\
			nnU-Net & $\checkmark$ & $\checkmark$ & \textbf{72.4} & \textbf{83.8} & \textbf{64.7} & \textbf{78.1} \\ \midrule
			SAM &  &  & 68.4 & 81.5 & 63.4 & 77.2 \\
			SAM & $\checkmark$ &  & 71.0 & 83.5 & 64.2 & 78.5 \\
			SAM &  & $\checkmark$ & 70.2 & 84.0 & 63.9 & 78.7 \\
			SAM & $\checkmark$ & $\checkmark$ & \textbf{73.1} & \textbf{85.3} & \textbf{64.8} & \textbf{79.6} \\ \bottomrule
		\end{tabular}%
	}
\end{table}

\noindent\textbf{Degeneration Avoidance in TADT.} 
A na\"ive learnable threshold often collapses into trivial bias shifting, where the network simply offsets both the logits and the threshold without refining the actual decision boundary. 
Table~\ref{tab:tadt_degen_xcad_crack500} verifies our explicit countermeasures against this degradation by detailing each removed component. 
The \textit{na\"ive learnable threshold} predicts a scalar threshold optimized only by standard pixel-wise loss, without any topology-aware surrogate, and thus yields the \emph{worst overall} performance across datasets and metrics. 
Notably, it may show a slightly higher clDice on XCAD compared to \textit{w/o clDice-on-surrogate}; this does not indicate better segmentation, but rather reflects a degenerate behavior where an overly permissive (lower) threshold produces thicker or over-connected predictions that \emph{appear} more continuous, while simultaneously introducing more false positives and degrading IoU. 
The \textit{w/o clDice-on-surrogate} variant introduces the differentiable soft-threshold but applies the clDice loss only on fixed 0.5-thresholded logits rather than the surrogate $b_k$, demonstrating that the threshold itself must be explicitly guided by structural connectivity. 
The \textit{w/o consistency loss} variant removes the local perturbation stability constraint $\mathcal{L}_{consist}$, making the dynamic decision boundary more vulnerable to localized background noise. 
Finally, the \textit{w/o logit centering} variant predicts the threshold directly from raw logits without subtracting the spatial mean $\mu(z_k)$, allowing the network to cheat by shifting the global feature distribution. 
Integrating all these constraints (\textbf{Full TADT}) effectively steers optimization towards a robust, topology-preserving cutting plane.
\begin{table}[htpb]
	\centering
	\caption{\textbf{Degeneration-avoidance ablation of TADT (instantiated in PEPA SAM).} Removing any countermeasure causes a regression towards the na\"ive learnable threshold.}
	\label{tab:tadt_degen_xcad_crack500}
	\resizebox{0.64\columnwidth}{!}{%
		\begin{tabular}{@{}lcccc@{}}
			\toprule
			\multicolumn{1}{c}{\multirow{2}{*}{Variant}} & \multicolumn{2}{c}{XCAD} & \multicolumn{2}{c}{Crack500} \\ \cmidrule(l){2-5} 
			& IoU $\uparrow$ & clDice $\uparrow$ & IoU $\uparrow$ & clDice $\uparrow$ \\ \midrule
			\textbf{Full TADT} & \textbf{73.1} & \textbf{85.3} & \textbf{64.8} & \textbf{79.6} \\
			w/o logit centering & 72.8 & 84.9 & 64.6 & 79.5 \\
			w/o consistency loss & 72.7 & 84.8 & 64.6 & 79.3 \\
			w/o clDice-on-surrogate & 71.2 & 82.8 & 64.5 & 78.9 \\
			na\"ive learnable threshold & 70.9 & 83.4 & 64.2 & 78.5 \\ \bottomrule
		\end{tabular}%
	}
\end{table}
\subsubsection{Complexity and Efficiency Analysis}
A practical post-encoder adapter must avoid introducing prohibitive computational overhead to the foundation model. 
Table~\ref{tab:tradeoff_avg} compares our approach against recent advanced upsampling operators. 
While attention-based methods like AnyUp~\cite{wimmer2025anyup} achieve strong metrics, they incur substantial parameter and latency penalties due to exhaustive patch-wise attention computations. 
In contrast, TCSU delivers superior topological accuracy with highly efficient feature aggregation by sampling strictly along 1D continuous chains. 

When fully equipped with PEPA (TCSU + TADT), the adapter adds merely 0.26M parameters and 0.22G FLOPs. 
We evaluated the end-to-end inference latency on a single NVIDIA H200 GPU with an input resolution of 1024$\times$1024 (batch size $=1$). 
Because the frozen foundation encoder dominates the overall runtime, PEPA introduces only a marginal end-to-end overhead (increasing latency from 99.0 ms to 103.0 ms, corresponding to a slight drop from 10.1 to 9.7 FPS). 
This confirms that PEPA remains a lightweight and deployment-friendly enhancement for structure-preserving segmentation.
\begin{table}[htpb]
	\centering
	\caption{\textbf{Accuracy--efficiency trade-off averaged across XCAD and Crack500 datasets.} 
		Extra Params/FLOPs denote the incremental model cost introduced by each method (decoder-side for CARAFE/DySample/AnyUp/TCSU/PEPA; encoder-side for LoRA). 
		End-to-end inference Time and FPS are measured on a single NVIDIA H200 GPU with a 1024$\times$1024 input (batch size $=1$).}
	\label{tab:tradeoff_avg}
	\resizebox{\columnwidth}{!}{%
		\begin{tabular}{@{}lcccccc@{}}
			\toprule
			Method & Avg IoU $\uparrow$ & Avg clDice $\uparrow$ & Extra Params (M)$\downarrow$ & Extra FLOPs (G)$\downarrow$ & Time (ms)$\downarrow$ & FPS$\uparrow$ \\ \midrule
			SAM (ViT-B) & 65.9 & 79.4 & +0.00 & +0.00 & \textbf{99.0} & \textbf{10.1} \\
			SAM + LoRA@Encoder & 67.2 & 80.9 & +0.26 & +0.05 & 101.0 & 9.9 \\ \midrule
			CARAFE~\cite{wang2019carafe} & 66.3 & 80.1 & +0.08 & +0.32 & 106.0 & 9.4 \\
			DySample~\cite{liu2023learning} & 66.2 & 80.3 & $-$0.03 & +0.06 & 100.0 & 10.0 \\
			AnyUp~\cite{wimmer2025anyup} & 66.5 & 80.6 & +0.73 & +0.58 & 112.0 & 8.9 \\ \midrule
			TCSU (Ours) & 67.6 & 81.6 & +0.21 & +0.18 & 102.0 & 9.8 \\
			\textbf{PEPA (TCSU+TADT)} & \textbf{68.9} & \textbf{82.5} & +0.26 & +0.22 & 103.0 & 9.7 \\ \bottomrule
		\end{tabular}%
	}
\end{table}
\subsubsection{Comparison with Encoder-side PEFT (LoRA)}
To further verify that PEPA's gains are not merely due to adding trainable parameters, we include an encoder-side PEFT baseline using Low-Rank Adaptation (LoRA) on the frozen ViT-B image encoder. 
Following common practice, we attach rank-4 LoRA modules to the attention projections (QKV in all transformer blocks and the output projection in the last six blocks), resulting in $\sim$0.26M additional trainable parameters---matched to the parameter overhead of PEPA.
All other settings are kept identical to our PEPA SAM experiments, including the SAM-HQ training protocol and the strict GT-box evaluation at test time. 
As summarized in Table~\ref{tab:tradeoff_avg}, under the same parameter budget, PEPA yields consistently larger gains on topology (clDice) while maintaining comparable end-to-end efficiency, supporting our choice of a post-encoder plug-in for curvilinear structure preservation.

\section{Conclusion}
We presented PEPA, a lightweight post-encoder plug-in for 2D curvilinear segmentation pipelines with accessible decoder/head features and target, query, or class descriptors. PEPA combines TCSU for structure-aware reconstruction and TADT for adaptive differentiable binarization, improving fragile topology without modifying frozen encoders. Experiments on five medical and industrial benchmarks show consistent gains, especially in clDice, with only $\sim$0.26M additional parameters.

\section*{Acknowledgements}
The authors gratefully acknowledge the financial support from the Chongqing Science and Technology Bureau under the 2024 Key Project of Technology Innovation and Application Development, Research and Application of Precision Interactive Integrated Medical Service Technology (Grant No. CSTB2024TIAD-KPX0046), and the Major Project of Technology Innovation and Application Development, Key Technologies and Platform Development of Adaptive Multi-Task Large Medical Models for Intelligent Diagnosis and Treatment (Grant No. CSTB2025TIAD-STX0029).
\clearpage


%
%
\bibliographystyle{splncs04}
\bibliography{main}

\clearpage

\appendix

\section*{Supplementary Material Overview}
This supplementary material provides additional technical details, experimental protocols, and extended results that complement the main paper.

\begin{itemize}
	\item \textbf{Section A: Additional Related Work} (Sec.~\ref{sec:appendix_related_work}) --- additional discussion of VFM adapters and topology-aware optimization for curvilinear segmentation.
	\item \textbf{Section B: Extended Formulation of PEPA} (Sec.~\ref{sec:appendix_pepa}) --- detailed formulations and configurations of TCSU and TADT, together with an algorithmic summary.
	\item \textbf{Section C: Additional Experimental Settings} (Sec.~\ref{sec:supp_exp}) --- dataset descriptions, evaluation metrics, default hyper-parameters, implementation details (e.g., sampling operator $\mathcal{S}$), and the LoRA baseline configuration.
	\item \textbf{Section D: Additional Experimental Results} (Sec.~\ref{sec:supp_results}) --- prompt robustness, sensitivity analyses (e.g., $K$ sweep), and mechanism analyses for the adaptive threshold and dynamic length, followed by additional qualitative results.
\end{itemize}

\section{Additional Related Work}
\label{sec:appendix_related_work}
\subsection{Adapters for Vision Foundation Models and Topology Constraints}
Parameter-Efficient Fine-Tuning (PEFT) has become the standard paradigm for transferring massive VFMs (e.g., SAM~\cite{ke2023segment,ravi2024sam}, DINO series~\cite{oquab2023dinov2,simeoni2025dinov3}) to downstream tasks without full retraining. 
Methods like SAM-Adapter~\cite{chen2023sam}, SAM2-Adapter~\cite{chen2024sam2}, and VFM-Adapter~\cite{chen2025vfm} inject task-specific knowledge into frozen encoders through lightweight modules.
For curvilinear structures specifically, recent works such as VesSAM~\cite{fu2025vessam} and UCS~\cite{zhu2025ucs} design specialized multi-prompt generators and sparse adapters for complex vessel and universal curve segmentation. 
Concurrently, preserving the structural integrity of these targets requires topology-aware optimization. Beyond standard pixel-wise metrics, clDice~\cite{shit2021cldice} introduced a morphological skeleton-based loss to guarantee topology preservation.
Recent extensions include DTU-Net~\cite{lin2023dtu}, which learns topological similarity via a data-driven dual-network, and CAPE~\cite{esmaeilzadeh2025cape}, which enforces global connectivity by penalizing shortest-path disconnections. 
By positioning our framework at the post-encoder interface, we decouple it from the frozen VFM encoder and directly impose topological constraints on the differentiable binarization surrogate, bridging foundation semantics with micro-level geometric fidelity.

\section{Extended Formulation of PEPA}
\label{sec:appendix_pepa}

In this section, we provide detailed mathematical formulations and network configurations for the Post-Encoder Plug-in Adapter (PEPA).

\subsection{Detailed Formulation of TCSU}

\paragraph{Subpixel Initialization.}
For a target upsampling scale factor $s$ (e.g., $s=2$), each low-resolution lattice point $(x,y)$ introduces $s^2$ subpixel centers. We denote the subpixel center indexed by $m\in\{1,\dots,s^2\}$ as:
\begin{equation}
	\mathbf{p}^{0}_{m}(x,y) = (x,y) + \boldsymbol{\pi}_m,
\end{equation}
where $\boldsymbol{\pi}_m = (\Delta x_m^{0},\, \Delta y_m^{0})$ represents preset fractional offsets (e.g., $\pm \tfrac{1}{4}$ on each axis for $s=2$) aligned to the high-resolution grid.

\paragraph{Dynamic Length Prediction.}
We predict a target-specific snake length $L_k$ from the query embedding $\mathbf{e}_k$:
\begin{equation}
	\hat{L}_k = 1 + (K-1)\cdot \sigma(\mathrm{MLP}_g(\mathbf{e}_k)), \qquad
	L_k = \mathrm{OddRound}(\hat{L}_k)\in\{1,3,\dots,K\},
\end{equation}
where $K$ is the maximal chain length and $\mathrm{OddRound}(\cdot)$ rounds to the nearest odd integer in $\{1,3,\dots,K\}$.

\paragraph{Target-Conditioned Deformation and Masking.}
The incremental bending offsets for the X-major ($\Delta y^{x}$) and Y-major ($\Delta x^{y}$) snakes are generated using a shared-plus-refinement architecture. For instance, the X-major offset before masking is defined as:
\begin{equation}
	f_{x}(\mathbf{F},\mathbf{e}_k;m) = f_{x,\text{sh}}(\mathbf{F};m) + \eta_k \cdot f_{x,\text{tg}}(\mathbf{F},\mathbf{e}_k;m), \qquad \eta_k=\sigma(\mathrm{MLP}_\eta(\mathbf{e}_k)),
\end{equation}
where $f_{x,\text{sh}}$ and $f_{x,\text{tg}}$ are lightweight conv branches.
To keep the dynamic length coupling differentiable, we apply a smooth length-aware mask $\omega(i;L_k)$:
\begin{equation}
	\Delta y^{x}_{k,m,i} \leftarrow \Delta y^{x}_{k,m,i}\cdot \omega(i;L_k), \qquad \Delta x^{y}_{k,m,i} \leftarrow \Delta x^{y}_{k,m,i}\cdot \omega(i;L_k),
\end{equation}
with $i\in[-c,c]$ and $c=\lfloor (K-1)/2 \rfloor$.
We instantiate $\omega(i;L)$ as a center-peaked sigmoid window:
\begin{equation}
	\omega(i;L)=\sigma\!\big(\gamma(\tfrac{L-1}{2}-|i|)\big),
\end{equation}
where $\gamma$ controls the softness (default $\gamma=2.0$).

\paragraph{Coordinate Accumulation.}
To preserve the structural adjacency prior of curvilinear objects, the masked increments are iteratively accumulated. For the X-major snake, which extends along the $x$-axis while bending in $y$, the accumulated vertical offsets are:
\begin{equation}
	\tilde{\Delta y}^{x}_{k,m}(0)=0,\qquad
	\tilde{\Delta y}^{x}_{k,m}(\pm i)=\tilde{\Delta y}^{x}_{k,m}(\pm (i-1)) + \Delta y^{x}_{k,m,\pm i},\quad i=1,\dots,c.
\end{equation}
The continuous sampling coordinates for the X-major snake are then:
\begin{equation}
	\mathbf{p}^{x}_{k,m,i}(x,y) = \Big(x + \Delta x^{0}_m + i,\; y + \Delta y^{0}_m + \tilde{\Delta y}^{x}_{k,m}(i)\Big), \qquad i\in[-c,c],
\end{equation}
and $\mathbf{p}^{y}_{k,m,i}$ is defined symmetrically for the Y-major snake.

\paragraph{Continuous Sampling.}
Given a sampling operator $\mathcal{S}$ (implemented by bilinear \texttt{grid\_sample}, see Sec.~\ref{sec:supp_hparams}), we obtain oriented sampled features:
\begin{equation}
	\mathbf{V}^{x}_{k,m,i} = \mathcal{S}\!\left(\mathbf{F}, \mathbf{p}^{x}_{k,m,i}\right),\qquad
	\mathbf{V}^{y}_{k,m,i} = \mathcal{S}\!\left(\mathbf{F}, \mathbf{p}^{y}_{k,m,i}\right).
\end{equation}
We denote $\mathbf{V}^{x}_{k,m}=\{\mathbf{V}^{x}_{k,m,i}\}_{i=-c}^{c}$ and $\mathbf{V}^{y}_{k,m}=\{\mathbf{V}^{y}_{k,m,i}\}_{i=-c}^{c}$.

\paragraph{Oriented Aggregation and Fusion.}
Sampled features $\mathbf{V}^{x}_{k,m}$ and $\mathbf{V}^{y}_{k,m}$ are aggregated via lightweight oriented operators $\mathcal{A}_x$ and $\mathcal{A}_y$:
\begin{equation}
	\mathbf{u}^{x}_{k,m} = \mathcal{A}_x\!\left(\mathbf{V}^{x}_{k,m};\, L_k\right),\qquad \mathbf{u}^{y}_{k,m} = \mathcal{A}_y\!\left(\mathbf{V}^{y}_{k,m};\, L_k\right),
\end{equation}
where $\mathcal{A}_x,\mathcal{A}_y$ are 1D depthwise convolutions along the ordered chain dimension, and positions outside $L_k$ are softly suppressed by $\omega(i;L_k)$.
The subpixel responses are rearranged to spatial blocks (pixel-shuffle style) and fused alongside a bilinear shortcut $\mathbf{U}^{\text{bi}}_k$:
\begin{equation}
	\mathbf{U}_{k} = \varphi\Big(\mathbf{U}^{\text{bi}}_k \;\Vert\; \psi\big(\mathrm{Rearrange}(\{\mathbf{u}^{x}_{k,m}\}) \;\Vert\; \mathrm{Rearrange}(\{\mathbf{u}^{y}_{k,m}\})\big)\Big),
\end{equation}
where $\psi(\cdot)$ and $\varphi(\cdot)$ are lightweight conv fusions (default: $1{\times}1$ conv + GELU, then $3{\times}3$ conv).


\subsection{Detailed Formulation of TADT}
TADT aims to calibrate a \emph{target-specific} binarization threshold for each query embedding $\mathbf{e}_k$, so that thin structures can be separated from cluttered backgrounds without relying on a fixed $0.5$ cutoff.
To make the threshold learnable and topology-aware, we optimize all objectives on a differentiable surrogate $\mathbf{b}_k$ while explicitly preventing degenerate solutions.
Concretely, the following three designs work together: \textit{(i)} bounding $t_k$ for numerical stability, \textit{(ii)} centering logits and using a soft-threshold surrogate for end-to-end optimization, and \textit{(iii)} enforcing local stability via a consistency loss.

\paragraph{Threshold Bounding.}
To maintain numerical stability in the logit domain, the target-adaptive threshold $t_k$ is bounded within a fixed range $[t_{\min}, t_{\max}]$ (default $[-5,5]$):
\begin{equation}
	t_k = t_{\min} + (t_{\max}-t_{\min})\cdot \sigma\!\big(\tilde{h}_t(\mathbf{e}_k,\mathbf{r})\big),
\end{equation}
where $\mathbf{r} = \mathrm{Summ}(\mathbf{F})$ denotes global average pooled features and $\tilde{h}_t$ is a lightweight MLP.

\paragraph{Logit Centering and Surrogate Binarization.}
To avoid trivial bias shifting, we center the logit map $\mathbf{z}_k\in\mathbb{R}^{H\times W}$ (predicted for $\mathbf{e}_k$) by subtracting its spatial mean:
\begin{equation}
	\tilde{\mathbf{z}}_k(x,y)=\mathbf{z}_k(x,y)-\mu(\mathbf{z}_k),\qquad
	\mu(\mathbf{z}_k)=\frac{1}{HW}\sum_{x,y}\mathbf{z}_k(x,y).
\end{equation}
The differentiable binarization surrogate is defined as:
\begin{equation}
	\mathbf{b}_k(x,y)=\sigma\!\left(\alpha\big(\tilde{\mathbf{z}}_k(x,y)-t_k\big)\right),
\end{equation}
where $\alpha$ controls sharpness (default $\alpha=1.0$).

\paragraph{Local Stability Optimization.}
To prevent the decision boundary from becoming overly sensitive to noise, we compute two perturbed surrogates:
\begin{equation}
	\mathbf{b}_k^{\pm}(x,y) = \sigma\!\left(\alpha\big(\tilde{\mathbf{z}}_k(x,y)-(t_k\pm\Delta)\big)\right),
\end{equation}
where $\Delta=0.5$. The consistency loss $\mathcal{L}_{\text{consist}}$ is computed as soft Dice agreement:
\begin{equation}
	\mathcal{L}_{\text{consist}}(\mathbf{b}_k^{+},\mathbf{b}_k^{-})
	=1-\frac{2\langle \mathbf{b}_k^{+},\mathbf{b}_k^{-}\rangle+\epsilon}{\|\mathbf{b}_k^{+}\|_1+\|\mathbf{b}_k^{-}\|_1+\epsilon},
\end{equation}
with $\epsilon=10^{-6}$.

\noindent\textbf{Algorithmic summary.}
Alg.~\ref{alg:pepa} summarizes the end-to-end forward path of PEPA (TCSU+TADT) and the corresponding training objective used in our implementation.
\label{sec:appendix_algo}
\begin{algorithm}[h]
	\small
	\caption{PEPA forward and training objective (per query embedding $\mathbf{e}_k$).}
	\label{alg:pepa}
	\begin{algorithmic}[1]
		\REQUIRE Encoder feature $\mathbf{F}$, query embedding $\mathbf{e}_k$, logit map $\mathbf{z}_k$
		\STATE Predict $L_k \in \{1,3,\dots,K\}$ from $\mathbf{e}_k$; compute mask $\omega(\cdot;L_k)$
		\STATE Predict increments $\Delta y^{x}_{k,m,i}$, $\Delta x^{y}_{k,m,i}$; apply $\omega$; accumulate to get sampling coords $\mathbf{p}^x_{k,m,i}$, $\mathbf{p}^y_{k,m,i}$
		\STATE Sample $\mathbf{V}^x_{k,m,i}=\mathcal{S}(\mathbf{F},\mathbf{p}^x_{k,m,i})$, $\mathbf{V}^y_{k,m,i}=\mathcal{S}(\mathbf{F},\mathbf{p}^y_{k,m,i})$
		\STATE Aggregate $\mathbf{u}^x_{k,m}=\mathcal{A}_x(\mathbf{V}^x_{k,m};L_k)$, $\mathbf{u}^y_{k,m}=\mathcal{A}_y(\mathbf{V}^y_{k,m};L_k)$; fuse with bilinear shortcut to obtain upsampled feature $\mathbf{U}_k$
		\STATE Predict bounded threshold $t_k$ from $(\mathbf{e}_k,\mathrm{Summ}(\mathbf{F}))$; center logits $\tilde{\mathbf{z}}_k=\mathbf{z}_k-\mu(\mathbf{z}_k)$
		\STATE Compute surrogate $\mathbf{b}_k=\sigma(\alpha(\tilde{\mathbf{z}}_k-t_k))$ and perturbed surrogates $\mathbf{b}^{\pm}_k=\sigma(\alpha(\tilde{\mathbf{z}}_k-(t_k\pm\Delta)))$
		\STATE Compute $\mathcal{L}_{\text{total}}=\lambda_{\text{bce}}\mathcal{L}_{\text{bce}}+\lambda_{\text{dice}}\mathcal{L}_{\text{dice}}+\lambda_{\text{cl}}(1-\mathrm{clDice})+\lambda_{\text{consist}}\mathcal{L}_{\text{consist}}$ on $\mathbf{b}_k$
	\end{algorithmic}
\end{algorithm}

\section{Additional Experimental Settings}
\label{sec:supp_exp}

\subsection{Datasets and Evaluation Protocols}
\label{sec:supp_datasets}
We evaluate PEPA on five public curvilinear segmentation benchmarks covering retinal vasculature, coronary angiography, and pavement cracks.

\begin{itemize}
	\item \textbf{DRIVE~\cite{fraz2012ensemble} (retinal fundus vessels).}
	This dataset was created for retinal vessel extraction in a screening setting in the Netherlands.
	It contains 40 color fundus photographs (single-field, $45^\circ$ field-of-view), captured with a Canon CR5 non-mydriatic 3CCD fundus camera.
	Vessel masks are provided as manual annotations by experts (with a second annotation available for part of the test set in common use).
	
	\item \textbf{CHASEDB1~\cite{fraz2012ensemble} (retinal fundus vessels in children).}
	CHASEDB1 comes from the \emph{Child Heart and Health Study in England (CHASE)}, a school-based cardiovascular health study.
	Retinal images were recorded in the field using a hand-held Nidek NM-200D fundus camera (around $30^\circ$ field-of-view), with two sets of expert vessel annotations provided for reference.
	The dataset contains 28 images (both eyes from 14 children) and is commonly used to test robustness on thinner vessels and higher-resolution fundus imagery.
	
	\item \textbf{CHUAC~\cite{cervantes2019automatic} (coronary angiography).}
	CHUAC is a small public benchmark of invasive X-ray coronary angiography (XRCA) images, provided by the CHUAC Hemodynamics Unit.
	It contains 30 single-channel angiograms (originally reported at $189{\times}189$), with corresponding binary vessel masks delineated by an expert cardiologist; many works resize images/masks (e.g., to $512{\times}512$) for model compatibility.
	
	\item \textbf{XCAD~\cite{ma2021self} (coronary angiography during intervention).}
	XCAD was built to facilitate coronary artery segmentation research and contains angiograms acquired during stent placement.
	Images were obtained using a General Electric Innova IGS 520 system and provided as single-channel frames (commonly $512{\times}512$).
	The released benchmark includes 1621 training angiograms paired with mask frames and 126 independent test angiograms with vessel masks annotated by experienced radiologists.
	
	\item \textbf{Crack500~\cite{yang2019feature} (pavement cracks).}
	Crack500 is a real-world pavement crack benchmark collected around the premises of Temple University using mobile-phone imagery.
	It contains 500 high-resolution RGB pavement images (commonly reported around $3264{\times}2448$), with pixel-level crack annotations.
	The dataset is widely used to evaluate thin, low-contrast crack patterns under cluttered road textures.
\end{itemize}

\noindent\textbf{Evaluation.}
Let $P$ and $G$ denote the predicted and ground-truth binary masks, respectively.
We report region overlap by Intersection-over-Union (IoU):
\begin{equation}
	\mathrm{IoU}(P,G)=\frac{|P\cap G|}{|P\cup G|}.
\end{equation}
To assess topological connectivity, we use centerline Dice (clDice).
Let $S_P=\mathrm{Skel}(P)$ and $S_G=\mathrm{Skel}(G)$ be the morphological skeletons (centerlines) of the prediction and ground truth.
We define topology precision and topology sensitivity as:
\begin{equation}
	T_{\mathrm{prec}}=\frac{|S_P\cap G|}{|S_P|+\epsilon},\qquad
	T_{\mathrm{sens}}=\frac{|S_G\cap P|}{|S_G|+\epsilon},
\end{equation}
and compute
\begin{equation}
	\mathrm{clDice}(P,G)=\frac{2\,T_{\mathrm{prec}}\,T_{\mathrm{sens}}}{T_{\mathrm{prec}}+T_{\mathrm{sens}}+\epsilon},
\end{equation}
where $\epsilon$ is a small constant for numerical stability.

\subsection{Hyper-parameters and Defaults}
\label{sec:supp_hparams}
Table~\ref{tab:supp_hparams} summarizes the default hyper-parameters used in our experiments.

\begin{table}[t]
	\centering
	\caption{\textbf{Default hyper-parameters in PEPA.}}
	\label{tab:supp_hparams}
	\resizebox{0.98\columnwidth}{!}{%
		\begin{tabular}{@{}ll@{}}
			\toprule
			\textbf{Component} & \textbf{Default value} \\
			\midrule
			\multicolumn{2}{l}{\textbf{TCSU (snake upsampling)}}\\
			Upsampling scale $s$ & $2$ (PEPA SAM instantiation) \\
			Max chain length $K$ & $9$ \\
			Length set $\mathcal{L}$ & $\{1,3,5,7,9\}$ \\
			Mask softness $\gamma$ in $\omega(i;L)$ & $2.0$ \\
			Oriented aggregation $\mathcal{A}_x/\mathcal{A}_y$ & 1D depthwise conv, kernel size $K$, zero padding \\
			Fusion $\psi(\cdot)$, $\varphi(\cdot)$ & $1{\times}1$ conv+GELU, then $3{\times}3$ conv \\
			\addlinespace[0.3em]
			\multicolumn{2}{l}{\textbf{TADT (adaptive thresholding)}}\\
			Threshold range $[t_{\min},t_{\max}]$ & $[-5,5]$ \\
			Soft-threshold sharpness $\alpha$ & $1.0$ \\
			Perturbation $\Delta$ (logit domain) & $0.5$ \\
			Dice epsilon $\epsilon$ & $10^{-6}$ \\
			Loss weights $(\lambda_{\text{bce}},\lambda_{\text{dice}},\lambda_{\text{cl}},\lambda_{\text{consist}})$ & $(1.0,\,1.0,\,1.0,\,0.2)$ \\
			\addlinespace[0.3em]
			\multicolumn{2}{l}{\textbf{Training / efficiency measurement}}\\
			Optimizer / scheduler & AdamW + cosine annealing \\
			Learning rate / weight decay & $1\times 10^{-4}$ / $1\times 10^{-4}$ \\
			Epochs / batch size (training) & $100$ / $8$ \\
			Input resolution & $1024{\times}1024$ \\
			Latency measurement & single forward, batch size $1$, end-to-end on H200 \\
			\bottomrule
	\end{tabular}}
\end{table}

We implement $\mathcal{S}$ using bilinear \texttt{grid\_sample}. 
Given a continuous coordinate $\mathbf{p}=(x,y)$ in feature-map pixel coordinates, we convert it to normalized grid coordinates and apply bilinear interpolation.
We use \texttt{align\_corners=False} and \texttt{padding\_mode=zeros} by default, so samples outside the feature boundary are treated as zeros.

\subsection{Encoder-side PEFT Baseline (LoRA) Details}
\label{sec:supp_lora}
For the parameter-matched encoder-side PEFT baseline, we apply LoRA to the frozen ViT-B image encoder with rank $r=4$: LoRA is attached to the attention QKV projections in all transformer blocks and to the output projection in the last six blocks, yielding $\sim$0.26M additional trainable parameters.
All other settings follow the main paper (same prompt protocol and training schedule). We train LoRA parameters together with the mask decoder while keeping the foundation encoder weights frozen.
Extra FLOPs are reported as an incremental estimate relative to the baseline forward pass.

\section{Additional Experimental Results}
\label{sec:supp_results}
\subsection{Prompt Robustness}

\begin{table}[h]
	\centering
	\caption{\textbf{Prompt robustness (average over XCAD and Crack500).} All prompts are derived from GT masks. \textit{Points} denote positive foreground clicks sampled from the GT region.}
	\label{tab:prompt_robustness}
	\resizebox{0.85\columnwidth}{!}{%
		\begin{tabular}{@{}lcccc@{}}
			\toprule
			\multirow{2}{*}{Prompt type (GT-derived)} & \multicolumn{2}{c}{SAM (ViT-B)} & \multicolumn{2}{c}{SAM + PEPA} \\ \cmidrule(l){2-5} 
			& Avg IoU $\uparrow$ & Avg clDice $\uparrow$ & Avg IoU $\uparrow$ & Avg clDice $\uparrow$ \\ \midrule
			Box (GT) & 65.9 & 79.4 & 68.9 & 82.5 \\
			1 positive point (GT) & 66.0 & 79.2 & 68.9 & 82.4 \\
			3 positive points (GT) & 65.8 & 80.1 & 69.6 & 83.3 \\
			Box (GT) + 1 point (GT) & 66.2 & 79.6 & 69.3 & 82.9 \\ \bottomrule
		\end{tabular}%
	}
\end{table}

Table~\ref{tab:prompt_robustness} reports PEPA SAM under different GT-derived prompt types on XCAD and Crack500.
Overall, PEPA yields consistent gains across all prompt configurations, and multi-point prompts (e.g., 3 points) typically provide stronger guidance than a single click, leading to better performance in both IoU and clDice.

\subsection{Sensitivity to Maximum Chain Length $K$}

\begin{figure}[h]
	\centering
	\includegraphics[width=\textwidth]{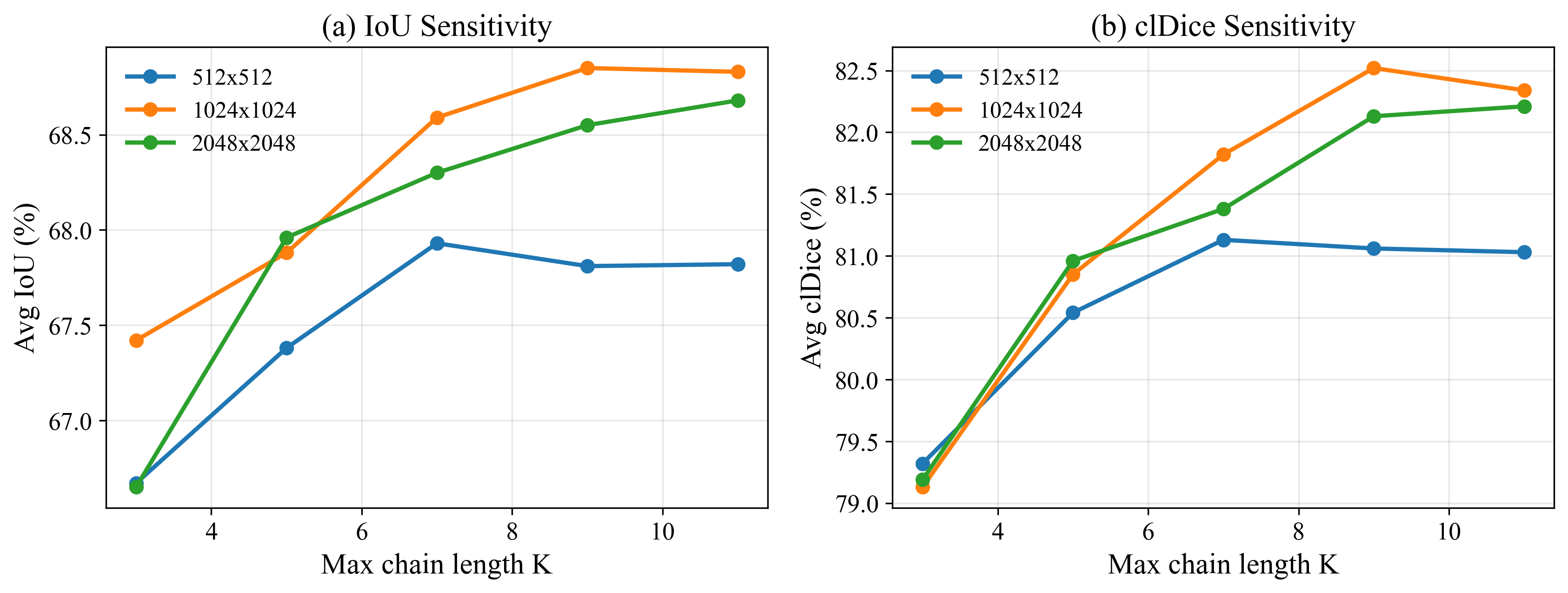}
	\caption{\textbf{Sensitivity to the maximal chain length $K$ under different input resolutions.}
		(a) IoU sensitivity and (b) clDice sensitivity as $K$ varies.
		The optimal $K$ shifts with resolution, suggesting that longer chains are more useful when finer structures are represented at higher spatial scales.}
	\label{fig:k_sweep}
\end{figure}

Fig.~\ref{fig:k_sweep} studies how the maximal snake chain length $K$ affects performance under different input resolutions.
We observe a clear resolution-dependent trend: using a larger $K$ becomes increasingly beneficial at higher resolutions, while overly large $K$ may provide limited gains (or slight saturation) at lower resolutions.
In particular, $K{=}9$ achieves the best trade-off around $1024{\times}1024$, consistent with our default setting in the main paper.

\subsection{Mechanism Analysis of TADT and TCSU}

\begin{figure}[h]
	\centering
	\includegraphics[width=\textwidth]{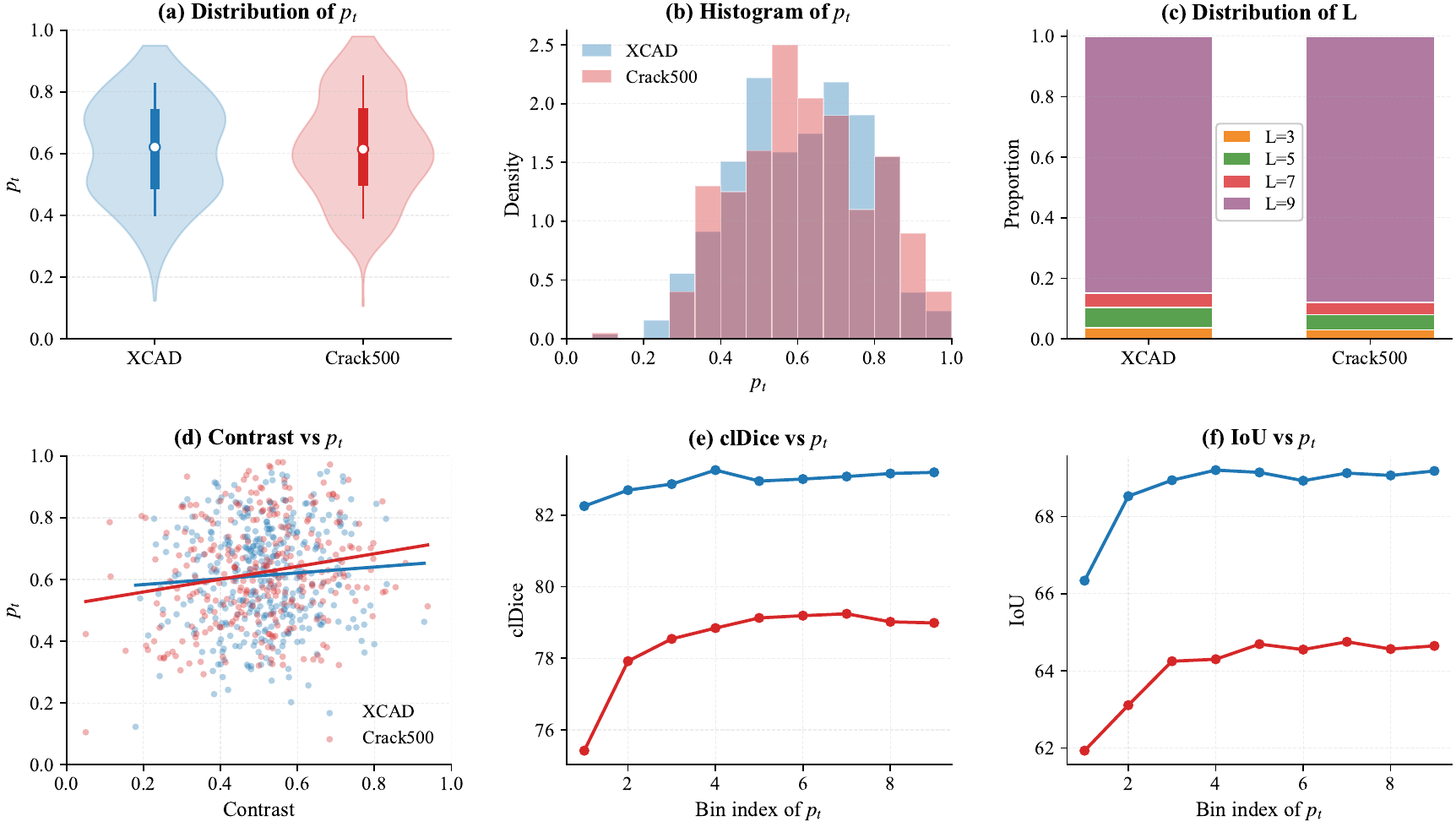}
	\caption{\textbf{Mechanism analysis of the adaptive threshold and dynamic length (XCAD and Crack500).}
		(a) Violin plot of the probability-domain threshold $p_t$.
		(b) Histogram of $p_t$.
		(c) Distribution of predicted dynamic length $L$.
		(d) Contrast proxy vs.\ $p_t$ with fitted trends.
		(e) clDice vs.\ binned $p_t$.
		(f) IoU vs.\ binned $p_t$.}
	\label{fig:mech}
\end{figure}
Fig.~\ref{fig:mech} provides additional analyses to help interpret the behavior of the learned adaptive threshold and dynamic length.
In our implementation, TADT predicts a logit-domain threshold $t_k\in[t_{\min},t_{\max}]$ for each query embedding $\mathbf{e}_k$, and binarization is performed via the soft-threshold surrogate $\mathbf{b}_k(x,y)=\sigma\!\big(\alpha(\tilde{\mathbf{z}}_k(x,y)-t_k)\big)$.
For readability, we visualize thresholds in the probability domain by mapping $t_k$ to
\begin{equation}
	p_t=\sigma(t_k)\in(0,1),
\end{equation}
where smaller $p_t$ corresponds to a more permissive decision boundary (i.e., more pixels tend to be activated as foreground), while larger $p_t$ yields a stricter boundary.

Across XCAD and Crack500, the predicted $p_t$ exhibits non-trivial, dataset-dependent distributions (Fig.~\ref{fig:mech}a--b), suggesting that TADT adapts the decision boundary to different imaging conditions.
We further show that the predicted dynamic length $L_k$ is highly skewed toward large values (Fig.~\ref{fig:mech}c), indicating that long-range oriented aggregation is frequently selected in TCSU.
To relate thresholding behavior to image difficulty, we compute a simple \emph{contrast proxy} per sample based on GT foreground/background separation:
\begin{equation}
	\mathrm{Contrast}=\frac{\big|\mu_{\mathrm{fg}}-\mu_{\mathrm{bg}}\big|}{\sigma_{\mathrm{bg}}+\epsilon},
\end{equation}
where $\mu_{\mathrm{fg}}$ is the mean intensity over GT foreground pixels, $\mu_{\mathrm{bg}}$ and $\sigma_{\mathrm{bg}}$ are the mean and standard deviation over background pixels (computed within the GT box region to reduce the influence of unrelated areas), and $\epsilon$ is a small constant.
As shown in Fig.~\ref{fig:mech}d, $p_t$ correlates positively with this contrast proxy: easier (higher-contrast) cases tend to adopt higher thresholds, whereas harder cases prefer lower thresholds to maintain structural continuity.

Finally, we study how performance varies with $p_t$ by grouping samples into 10 equal-width bins on $(0,1)$, i.e., $\{[0,0.1),[0.1,0.2),\dots,[0.9,1.0)\}$.
The \emph{bin index} in Fig.~\ref{fig:mech}e--f refers to the integer index of the interval that contains $p_t$ (from 1 to 10), and the curves report the average clDice/IoU within each bin.
The resulting trends indicate that varying $p_t$ is associated with systematic changes in clDice/IoU, supporting that TADT calibrates decision boundaries in a target-adaptive manner.

\section{Additional Qualitative Results}
\label{sec:supp_vis_fail}
\subsection{DRIVE}
\begin{figure}[h]
	\centering
	\includegraphics[width=\textwidth]{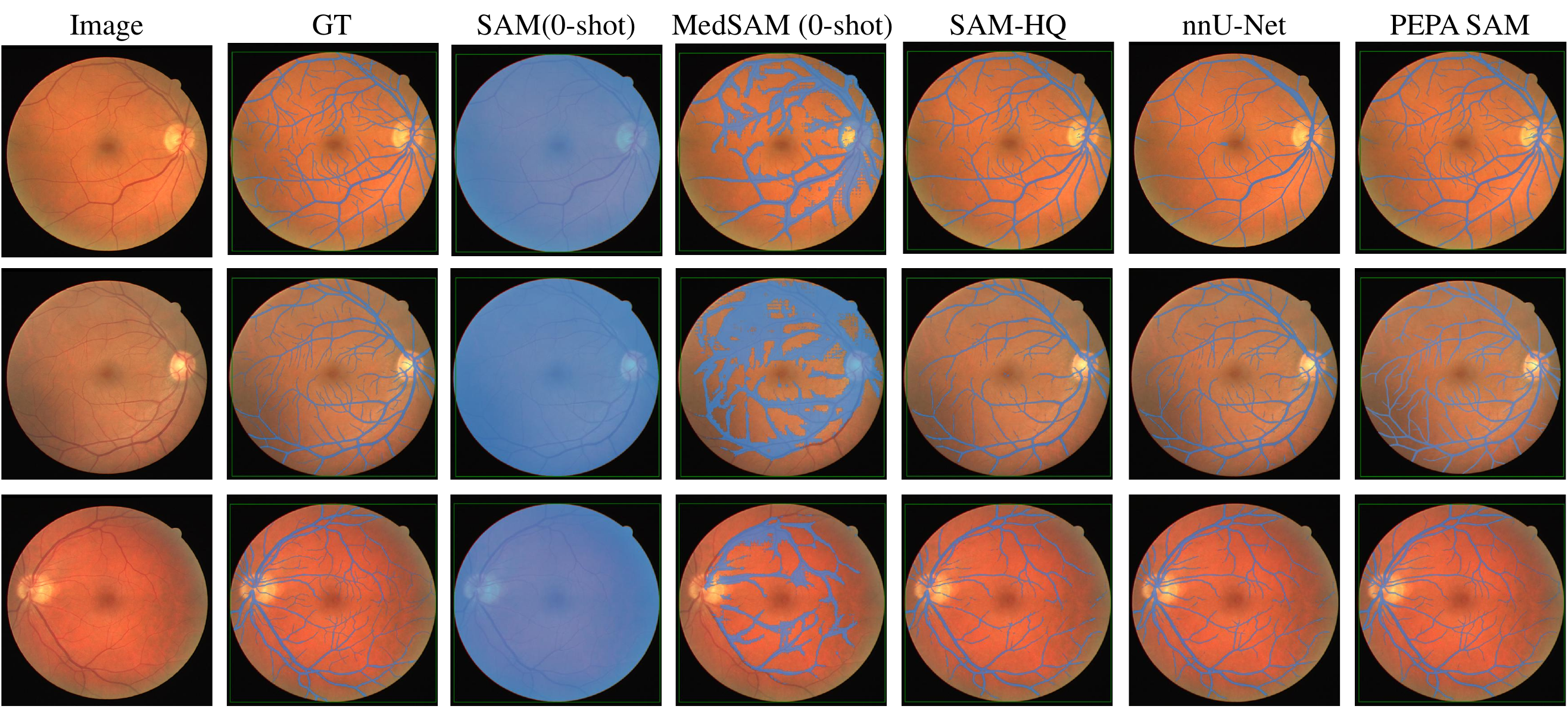}
	\caption{{Additional qualitative results on DRIVE.}
		From left to right: Image, GT, SAM (0-shot), MedSAM (0-shot), SAM-HQ, nnU-Net, and PEPA SAM.}
	\label{fig:supp_drive}
\end{figure}

Additional qualitative comparisons on DRIVE are shown in Fig.~\ref{fig:supp_drive}.

\subsection{CHASEDB1}
\begin{figure}[h]
	\centering
	\includegraphics[width=\textwidth]{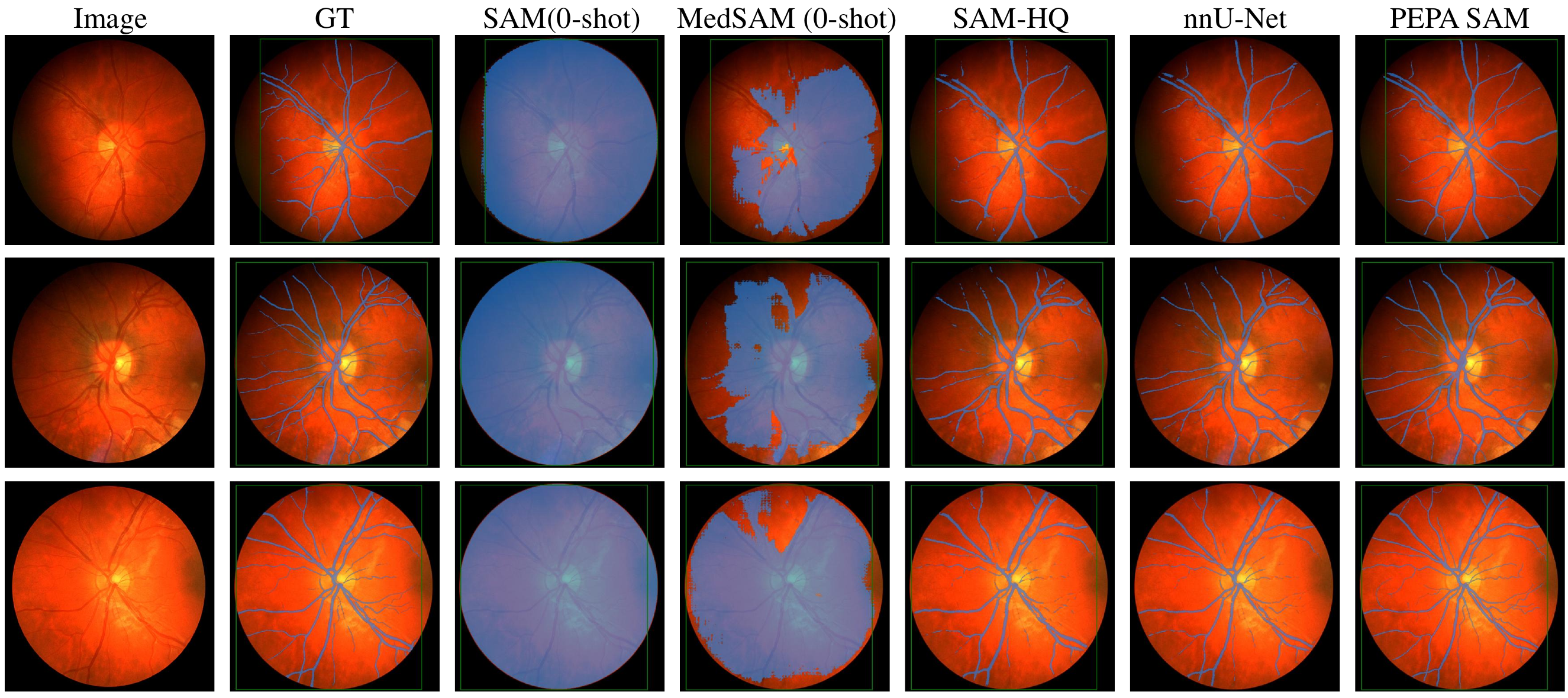}
	\caption{\textbf{Additional qualitative results on CHASEDB1.}
		From left to right: Image, GT, SAM (0-shot), MedSAM (0-shot), SAM-HQ, nnU-Net, and PEPA SAM.}
	\label{fig:supp_chasedb1}
\end{figure}

Additional examples on CHASEDB1 are provided in Fig.~\ref{fig:supp_chasedb1}.

\subsection{CHUAC}
\begin{figure}[h]
	\centering
	\includegraphics[width=\textwidth]{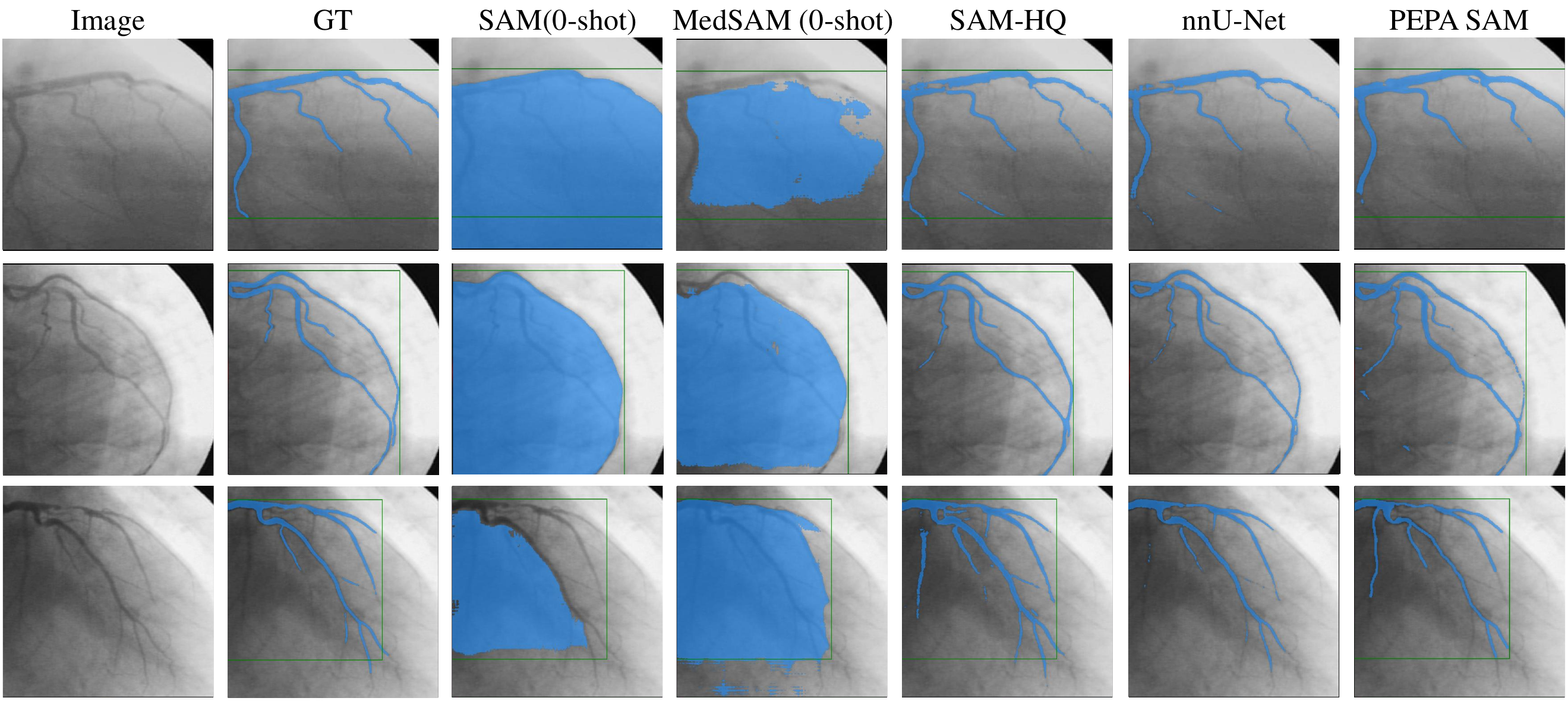}
	\caption{\textbf{Additional qualitative results on CHUAC.}
		From left to right: Image, GT, SAM (0-shot), MedSAM (0-shot), SAM-HQ, nnU-Net, and PEPA SAM.}
	\label{fig:supp_chuac}
\end{figure}

Additional examples on CHUAC are shown in Fig.~\ref{fig:supp_chuac}.
\vspace{-0.3em}

\subsection{XCAD}
\begin{figure}[h]
	\centering
	\includegraphics[width=\textwidth]{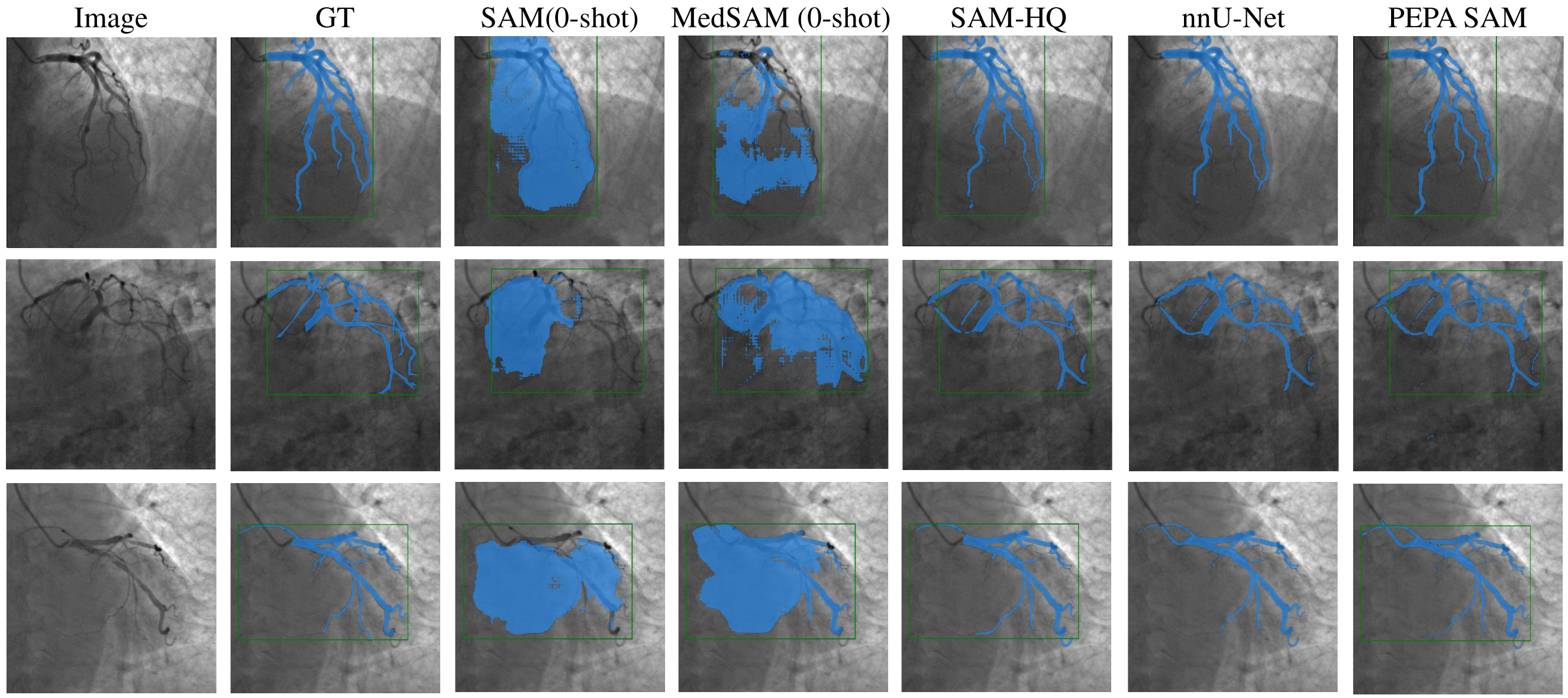}
	\caption{\textbf{Additional qualitative results on XCAD.}
		From left to right: Image, GT, SAM (0-shot), MedSAM (0-shot), SAM-HQ, nnU-Net, and PEPA SAM.}
	\label{fig:supp_xcad}
\end{figure}

Additional qualitative results on XCAD are provided in Fig.~\ref{fig:supp_xcad}.

\subsection{Crack500}
\begin{figure}[h]
	\centering
	\includegraphics[width=\textwidth]{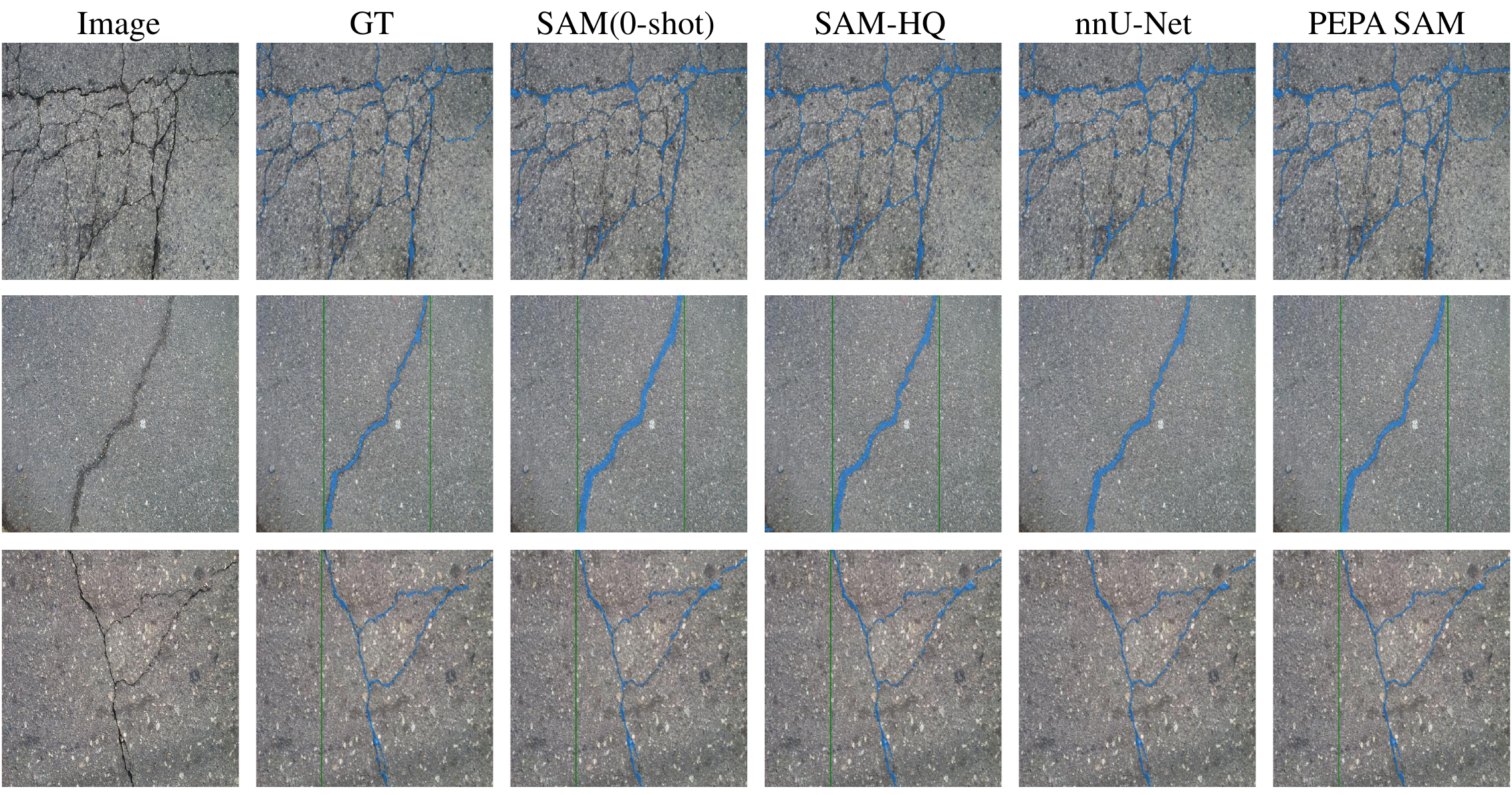}
	\caption{\textbf{Additional qualitative results on Crack500.}
		From left to right: Image, GT, SAM (0-shot), SAM-HQ, nnU-Net, and PEPA SAM.}
	\label{fig:supp_crack500}
\end{figure}

Additional crack segmentation examples on Crack500 are shown in Fig.~\ref{fig:supp_crack500}, including thin cracks under noisy pavement textures.

\end{document}